\documentclass[letterpaper]{article}

\usepackage{arxiv}
\usepackage{times}
\usepackage{helvet}
\usepackage{courier}
\usepackage[hyphens]{url}
\usepackage{graphicx}
\usepackage{natbib}
\usepackage{caption}
\usepackage{algorithm}
\usepackage{algorithmic}
\usepackage{color}
\usepackage{amsmath,amsfonts,bm}
\usepackage{booktabs}
\usepackage{newfloat}
\usepackage{listings}
\title{Can We Find Strong Lottery Tickets in Generative Models?}

\author {
    Sangyeop Yeo,\textsuperscript{\rm 1}
    Yoojin Jang,\textsuperscript{\rm 1}
    Jy-yong Sohn,\textsuperscript{\rm 2}
    Dongyoon Han,\textsuperscript{\rm 3}
    Jaejun Yoo\textsuperscript{\rm 1}\textsuperscript{*}
}

\affiliations {
    \textsuperscript{\rm 1} Laboratory of Advanced Imaging Technology (LAIT), Ulsan National Institute of Science and Technology (UNIST)\\
    \textsuperscript{\rm 2}University of Wisconsin-Madison\\
    \textsuperscript{\rm 3}Naver AI Lab\\
    sangyeop377@gmail.com,
    softjin@unist.ac.kr, jysohn1108@gmail.com,
    dongyoon.han@navercorp.com,
    jaejun.yoo@unist.ac.kr (*: corresponding author)
}

\begin{document}

\maketitle

\begin{abstract}
Yes. In this paper, we investigate  \emph{strong lottery tickets} in generative models, the subnetworks that achieve good generative performance without any weight update. Neural network pruning is considered the main cornerstone of model compression for reducing the costs of computation and memory. Unfortunately, pruning a generative model has not been extensively explored, and all existing pruning algorithms suffer from excessive weight-training costs, performance degradation, limited generalizability, or complicated training. To address these problems, we propose to find a strong lottery ticket via moment-matching scores. Our experimental results show that the discovered subnetwork can perform similarly or better than the trained dense model even when only 10\% of the weights remain. To the best of our knowledge, we are the first to show the existence of strong lottery tickets in generative models and provide an algorithm to find it stably. Our code and supplementary materials are publicly available.
\footnote{https://lait-cvlab.github.io/SLT-in-Generative-Models/}
\end{abstract}

\section{Introduction}

State-of-the-art generative models tend to use extremely large and complex structures for better performance~\citep{brock2018large, karras2019style, karras2020analyzing, ramesh2021zero, radford2021learning}. One downside of large models is the high computational costs for training, which limits their application to edge devices such as mobile environments.  
This naturally calls for the design of a new lightweight architecture or a new compression method in generative modeling.  

In this work, we focus on the model pruning techniques that fall into the latter category. Unlike discriminative models where various pruning techniques~\citep{lecun1989optimal,hassibi1992second,han2015learning,frankle2018lottery,ramanujan2020s,sreenivasan2022rare} have been actively studied, pruning generative models have not been extensively explored. Moreover, it has been found that na{\"i}ve application of existing pruning methods (developed for discriminative models) to generative models leads to performance degradation and/or unstable training~\citep{wang2020gan,li2021revisiting}.

\begin{figure}[t!]
\centering
\includegraphics[width=1.05\columnwidth]{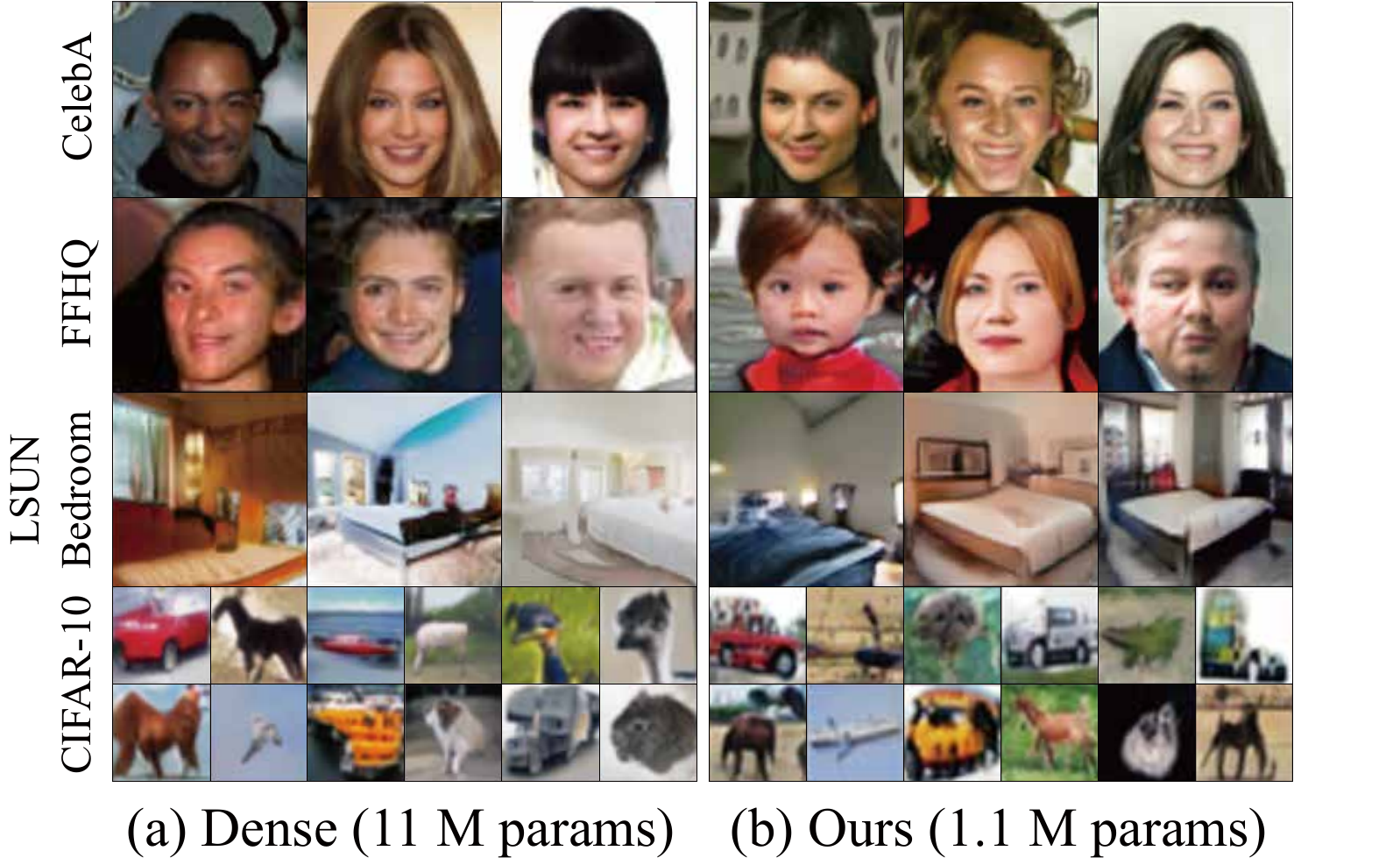}

\caption{\textbf{Visualization of dense models and our strong lottery tickets}. 
For various datasets including CelebA, FFHQ, LSUN Bedroom, and CIFAR-10, we compare the images generated by (a) a trained dense generative feature matching network (GFMN); (b) a subnetwork discovered by our method, which uses only 10\% of parameters within a randomly-initialized GFMN. Intriguingly, our method finds the strong lottery tickets which achieve comparable or even better generative performances over the baseline dense model qualitatively and quantitatively (Table~\ref{table:fid_4_datasets}).
}
\label{fig:teaser}
\end{figure}

Recently, several methods for pruning generative models have been proposed and showed that it is possible to obtain a lightweight model by following the ``train, prune, re-train" paradigm when tuned to generative models with special care~\citep{li2020gan, wang2020gan, liu2021content, li2021revisiting, tuli2021generative, hou2021slimmable}. For example, to overcome training instability, Hou \textit{et al.}~\citep{hou2021slimmable} introduced  multiple shared discriminators to train a slimmable generator that can flexibly change its capacity at runtime. Li \textit{et al.}~\citep{li2021revisiting} proposed a cooperative scheme between the generator and the discriminator to stabilize the compression during the adversarial training.

\begin{figure*}[ht!]

\centering
\includegraphics[width=0.81\paperwidth]{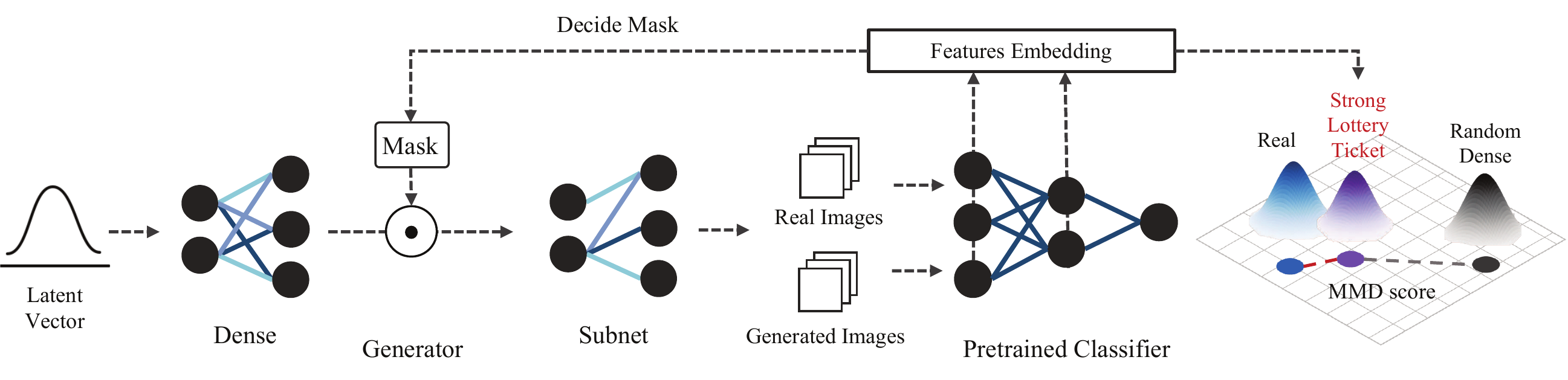}
\begin{tabular}{cc}
\end{tabular}
\caption{\textbf{A schematic overview of our method.} Our method finds a strong lottery ticket (SLT) via moment-matching scores. By exploiting the power of the pretrained classifier, our method assigns scores to the randomly initialized weights and finds a sparse mask so that the discovered subnetwork performs similarly or better than the trained dense generator.} 
\label{fig:overview}
\end{figure*}

However, because their basic strategy inevitably involves subtle balancing between training and pruning procedures, all existing methods suffer from excessive computational costs~\citep{liu2021content, tuli2021generative, hou2021slimmable}, performance degradation~\citep{wang2020gan, li2021revisiting, chen2021gans}, limited generalizability~\citep{li2020gan, hou2021slimmable}, or complicated training~\citep{liu2021content}. This, combined with the notorious instability of generative adversarial networks (GANs), which most methods target, makes it more challenging to develop a pruning method for generative models.

To address these problems, we propose to find \textit{strong lottery tickets} in generative models. A strong lottery ticket is a subnetwork at initialization (\textit{i.e.}, no weight update) that performs similarly or even better than its dense counterpart whose weights are trained. Here, we employ the edge-popup (EP) algorithm~\citep{ramanujan2020s}, which is the earliest method to find a strong lottery ticket in discriminative models. The EP algorithm selects a subnetwork mask based on the idea that one can ``score" the importance of each weight. Once such a score is assigned, one simply keeps the weights of high scores according to the desired target sparsity. 

Because the performance of the EP algorithm largely depends on the updated scores that serve as pruning criteria, it is essential to use a proper score function that gives a representative feature for pruning generative models. One may easily think of the adversarial loss, a commonly used criterion for training high-quality generators, but it is extremely unstable and hinders the search for appropriate scores.   
Instead, we propose to utilize a technique from statistical hypothesis testing known as maximum mean discrepancy (MMD)~\citep{gretton2006kernel,gretton2012kernel}, which leads to a simple moment-matching score using features extracted from a fixed, pretrained ConvNets~\citep{li2015generative, li2017mmd, binkowski2018demystifying, wang2018improving, santos2019learning, ren2021improving}.

By combining the EP algorithm with the moment-matching score, we propose a stable algorithm that finds a subnetwork with good generative performance in a very sparse regime. 
Note that our method can avoid the challenging problem of balancing between training and pruning procedures because it does not involve any weight update. In addition, thanks to the stable characteristic of the moment-matching score, our method can find a Strong Lottery Ticket (SLT) in generative models without bells and whistles. To the best of our knowledge, we are the first to show the existence of strong lottery tickets in generative models and provide an algorithm to find it stably. 
Our extensive experiments show that one can find a subnetwork of 10\% sparsity while maintaining the generative performance of its dense version (see Figure~\ref{fig:teaser}). More surprisingly, we find that our method can also be used to find a well-performing subnetwork in the pretrained generative models. This implies that one can scale down off-the-shelf generative models to have less memory consumption with comparable or even better performance. 
\paragraph{Main Contributions.}
Our contributions can be summarized as follows:
\begin{itemize}
    \item We show that there exist strong lottery tickets in generative models. By searching for strong lottery tickets via moment-matching scores, we avoid the joint optimization of pruning and training, which is complicated.
    \item We provide an algorithm that can stably find a good subnetwork in generative models; \textit{i.e.}, one can prune a randomly initialized generative model (without any weight updates) and find a sparse subnetwork that achieves comparable or better performance than the dense, fully trained counterpart. 
    \item We further find that our method can even improve pretrained generative models. Starting from a densely pretrained model, our method can produce its lighter and stronger counterpart in various experimental settings. 
\end{itemize}

\section{Method}
This section proposes a simple method to find a strong lottery ticket (SLT) in generative models. The schematic overview of our method is shown in Figure~\ref{fig:overview}. Here, we consider a neural network $G(\bm{z};\boldsymbol{\theta})$ with randomly initialized weights $\boldsymbol{\theta} \in \mathbb{R}^d$. We then aim at finding a strong lottery ticket: a mask $\bm{m} \in \{0,1\}^d$ which satisfies that the pruned network $G(\bm{z};\boldsymbol{\theta} \odot \bm{m})$ performs well on the generative task. 

\subsection{A Simple Algorithm for Finding Strong Lottery Tickets in Generative Models}
Edge-popup (EP) algorithm~\citep{ramanujan2020s} is the earliest method to find strong lottery tickets in randomly initialized discriminative networks. With a proper score function, we show that the EP algorithm can be successfully applied to generative models as well. In the EP algorithm, we first assign a random score $s_i$ for each weight $\theta_i$ where $\boldsymbol{\theta} = [\theta_1, \cdots, \theta_d]$. Suppose we want to remain $k\%$ of the weights. Then, at each forward path, we sort the score $s_i$ at each layer and assign $m_i = 1$ if $\lvert s_i \rvert$ is in the top $k\%$ within the corresponding layer, and assign $m_i = 0$ otherwise. In each backward path, we compute the loss of the network and update the score $s_i$ by using back-propagation. Here we use straight-through estimator~\citep{bengio2013estimating} to handle the indicator function that maps $s_i$ to $m_i$.

\subsection{Modeling a Stable Score via Moment-Matching}

Now we are pruning generative models, we need to devise a proper score-updating function instead of the cross-entropy loss used for discriminative models. To this end, we utilize a kernel maximum mean discrepancy (MMD)~\citep{gretton2006kernel, gretton2012kernel}, which is known to give a stable optimization for learning generative models~\citep{li2015generative, li2017mmd, binkowski2018demystifying, wang2018improving, santos2019learning, ren2021improving}.

Given two sets of real and fake samples $\{r_i\}^N_{i=1}$ and $\{f_i\}^M_{i=1}$, minimizing the MMD loss $\mathcal{L_{MMD}}$ can be interpreted as matching all moments of the model distribution to the empirical data distribution: 
\begin{equation}
\mathcal{L_{MMD}}= ||\frac{1}{N} \sum_{i=1}^N \phi(r_i) - \frac{1}{M}\sum_{j=1}^M \phi(f_j)||^2,
\label{eqmmd}
\end{equation}
where $\phi(\cdot)$ denotes a function that leads to matching high order moments. Ideally, $\phi(\cdot)$ must be calculated with infinite orders. To compute MMD efficiently, we rephrase the expression (\ref{eqmmd}) via kernel trick:
\begin{align}
\mathcal{L_{MMD}} =& 
\frac{1}{N^2} \sum_{i=1}^N\sum_{i'=1}^N \psi(r_i, r_{i'}) -
\frac{2}{NM} \sum_{i=1}^N\sum_{j=1}^M \psi(r_i, f_{j}) \nonumber\\ &+
\frac{1}{M^2} \sum_{j=1}^M\sum_{j'=1}^M \psi(f_j, f_{j'}),
\label{eqkernel}
\end{align}

where we use the pretrained VGG network as a fixed kernel $\psi$ and match the mean $\mu$ and covariance $\sigma$ of real and fake sample 
features in the VGG embedding space:
\begin{equation}
\mathcal{L_{MMD}}= \sum_{j=1}^L||\mu_{r}^j - \mu_{f}^j||^2 + ||\sigma_{r}^j - \sigma_{f}^j||^2.
\label{eqfm}
\end{equation}
We define $I_v$, $w_{uv}$, $\sigma$, and $\alpha$ as the input of node $v$, network parameter for node $u$ and node $v$, activation function, and learning rate, respectively. At time step $t$, the amount of changes in the score can be expressed as 
\begin{align}
\mathbf{s_{t+1, uv}}= \mathbf{s_{t, uv}} - \alpha\frac{\partial \mathcal{L_{MMD}} }{\partial I_v}w_{uv}\sigma(I_u).
\label{scoreupdate}
\end{align}
It is worth noting that our method uses the MMD loss for finding nodes of low importance, not for learning weights.

\section{Experimental Evaluation}
In this section, we provide empirical results on the proposed pruning method. First, we show that our method finds strong lottery tickets in randomly initialized generative models. 
Second, we show that our pruning method can be used to lighten pretrained generative models. Finally, we demonstrate that the strong lottery tickets in generative models found by our method are not fine-tunable to reach better performance, similar to the observation made for discriminative models~\citep{ramanujan2020s}.

\paragraph{Datasets.}
We use LSUN Bedroom~\citep{yu15lsun}, FFHQ~\citep{karras2019style}, CIFAR-10~\citep{krizhevsky2009learning}, CelebA~\citep{liu2015deep}, and BabyImageNet~\citep{kang2022studiogan} datasets. Image resolution is set to 64$\times$64 for every dataset.

\paragraph{Baselines.}
Following the setup of the generative feature matching network (GFMN)~\citep{santos2019learning}, we adopt the ResNet-based architecture as our default generator that serves as a dense model. We also use other off-the-shelf pretrained generative models (BigGAN, SNGAN) trained on BabyImageNet, whose weights are provided in the official StudioGAN ~\citep{kang2022studiogan} code\footnote{https://github.com/postech-cvlab/pytorch-studiogan}.  
The model setup is configured by the codebase implemented for reproducible GANs.

\begin{figure}[!t]
\centering
\includegraphics[width=\columnwidth ]{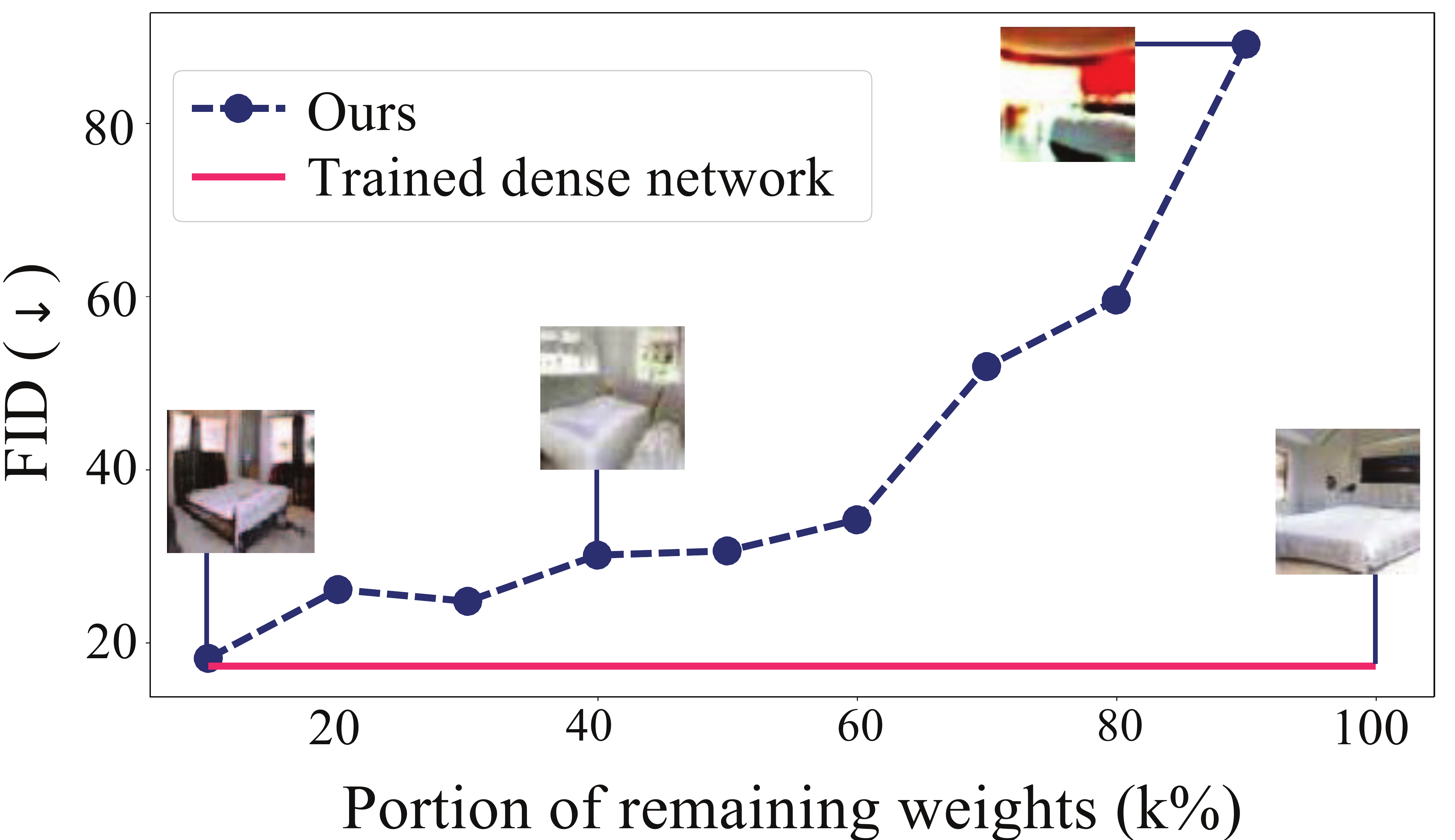}
\caption{ \textbf{Comparison of FID scores of the subnetworks and the trained dense network (GFMN; LSUN-Bedroom).} 
Recall that our method prunes a randomly initialized neural network without any weight update. Here, we visualize the FIDs for various $k$, which is the portion $(\%)$ of the remaining weights in the pruned subnetwork. 
}

\label{fig:varyingk}
\end{figure}

\paragraph{Evaluation metrics.} 
We evaluate the visual quality and the diversity of generated images with Fr\'echet Inception Distance (FID)~\citep{heusel2017gans}, Precision \& Recall~\citep{kynkaanniemi2019improved}, and Density \& Coverage~\citep{naeem2020reliable}, where we use InceptionV3 as the evaluation backbone model~\citep{szegedy2016rethinking}. Here, we use 10,000 samples of real and generated images, respectively. The details on evaluation metrics and protocols are further described in Supplementary Materials.

\subsection{Experiment 1: Can we find strong lottery tickets in generative models?}
To investigate this question, we need a dense model that serves as the reference. On the one hand, following the setup of GFMN~\citep{santos2019learning}, we train a ResNet-based generator using the MMD loss. Here, the loss is used for training model weights. 
On the other hand, we apply our method to find a subnetwork from the generator of the same architecture but with randomly initialized weights. Note that the same MMD loss is used here, but it just serves as a score function to find a subnetwork mask---the loss does not affect the weights. 
\begin{figure}[!t]
\centering
\includegraphics[width=\columnwidth ]{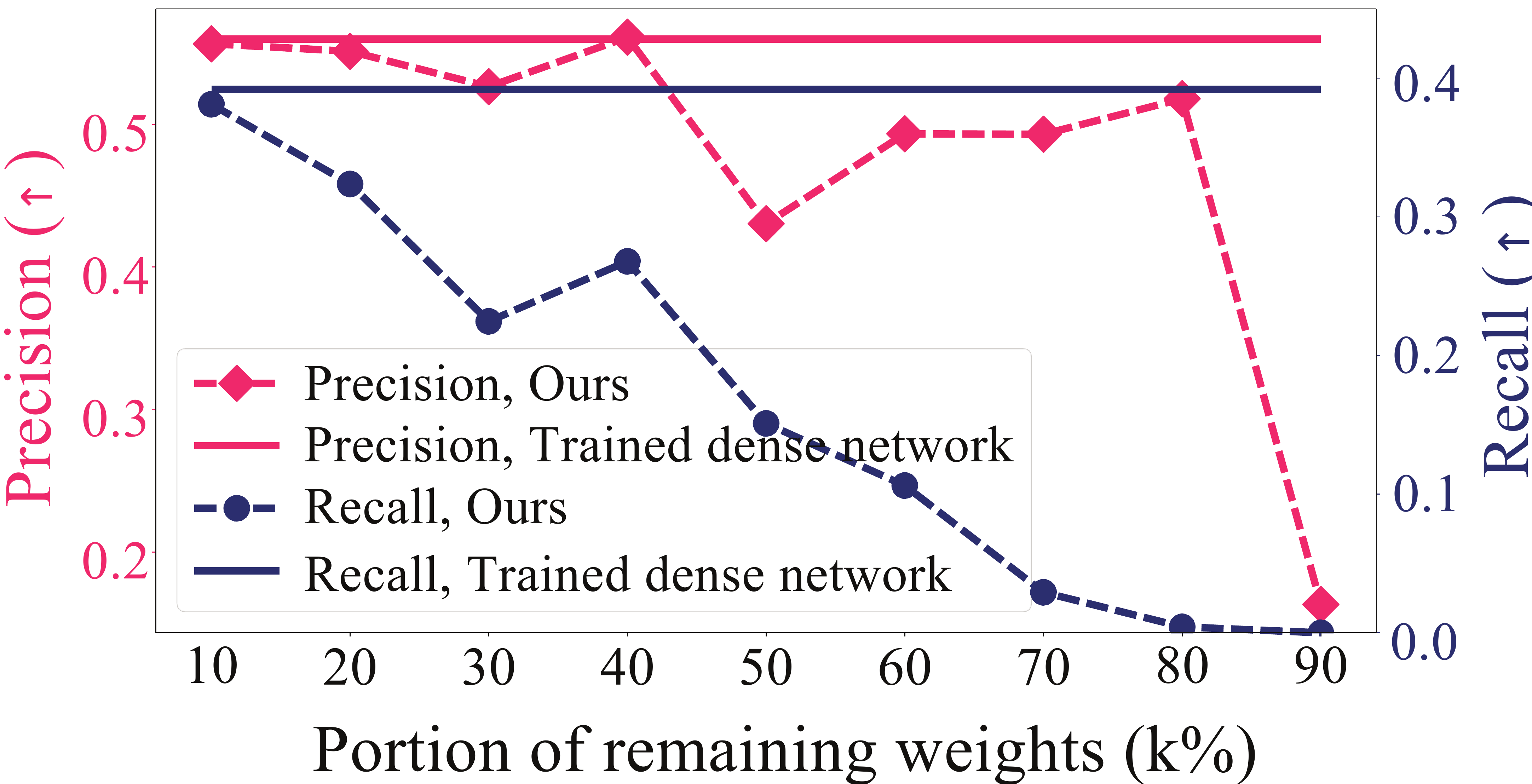}
\caption{ \textbf{Comparison of Precision \& Recall values of the subnetworks and the trained dense network (GFMN; LSUN-Bedroom).} The discovered SLT performs well in both Precision \& Recall. That is, SLT can generate ``various" images of ``good quality" without weight training.}

\label{fig:varyingk-pr}
\end{figure}
\begin{table}[t]
    \centering
    \fontsize{9}{10}
		 {
			\begin{tabular}{c|cccc}
				\toprule {\shortstack{}}
				& FFHQ & LSUN & CIFAR10 & CelebA
				\\
				\midrule
				Dense Network
				& 11.52
				& 17.32
				& 18.86
				& 9.52
				\\
				Ours ($k=10\%$) 
				& 13.32
				& 20.21
				& 15.06
				& 10.93
				\\
				\bottomrule
			\end{tabular}}%
    \caption{ 
    	\textbf{Comparison of FID values of trained dense models and strong lottery tickets (SLT).} SLT is found by our pruning method on a ResNet-based generator for various datasets. Smaller FID numbers indicate better performance. Here $k$ denotes the portion of remaining weights. This shows that we can obtain a decent generative model by pruning 90\% of weights in a randomly initialized neural network without any weight update.}
	\label{table:fid_4_datasets}
\end{table}
Figure~\ref{fig:varyingk} and Figure~\ref{fig:varyingk-pr} show how the generative performance of our pruned network changes as a function of $k$, the portion $(\%)$ of remaining weights in the subnetwork. 
When most of the random weights remain, \textit{e.g.}, $k=90\%$, the pruned network is almost identical to the untrained dense network and thus shows poor generative performance.
As $k$ decreases, one can see that the pruned network starts to generate realistic images and achieve its best FID values at around $k=10\%$. The similar trend is observed in Precision \& Recall (see Figure~\ref{fig:varyingk-pr}).

In Table~\ref{table:fid_4_datasets}, we compare the FID values of the trained dense models and the strong lottery tickets (SLT) that are obtained by our method (using $k=10\%$ of the weights).
This indicates that in a randomly initialized dense network, there is a subnetwork with similar or better performance than the trained dense model while having only 10\% of the total number of parameters in the dense network. Note that in the previous literature~\citep{ramanujan2020s}, \citet{ramanujan2020s} found the best performing SLT for discriminative networks in the $k=50\%$ region but failed to obtain SLT in a sparser region like $k=10\%$. 
In contrast, SLTs found by our method for generative models achieve their best performance when $k=10\%$. This difference of discriminate/generative models in the optimal sparsity regime for finding SLTs is an interesting observation, which can be further analyzed in future work.

\begin{figure}[t]
\centering
\includegraphics[width=1\columnwidth ]{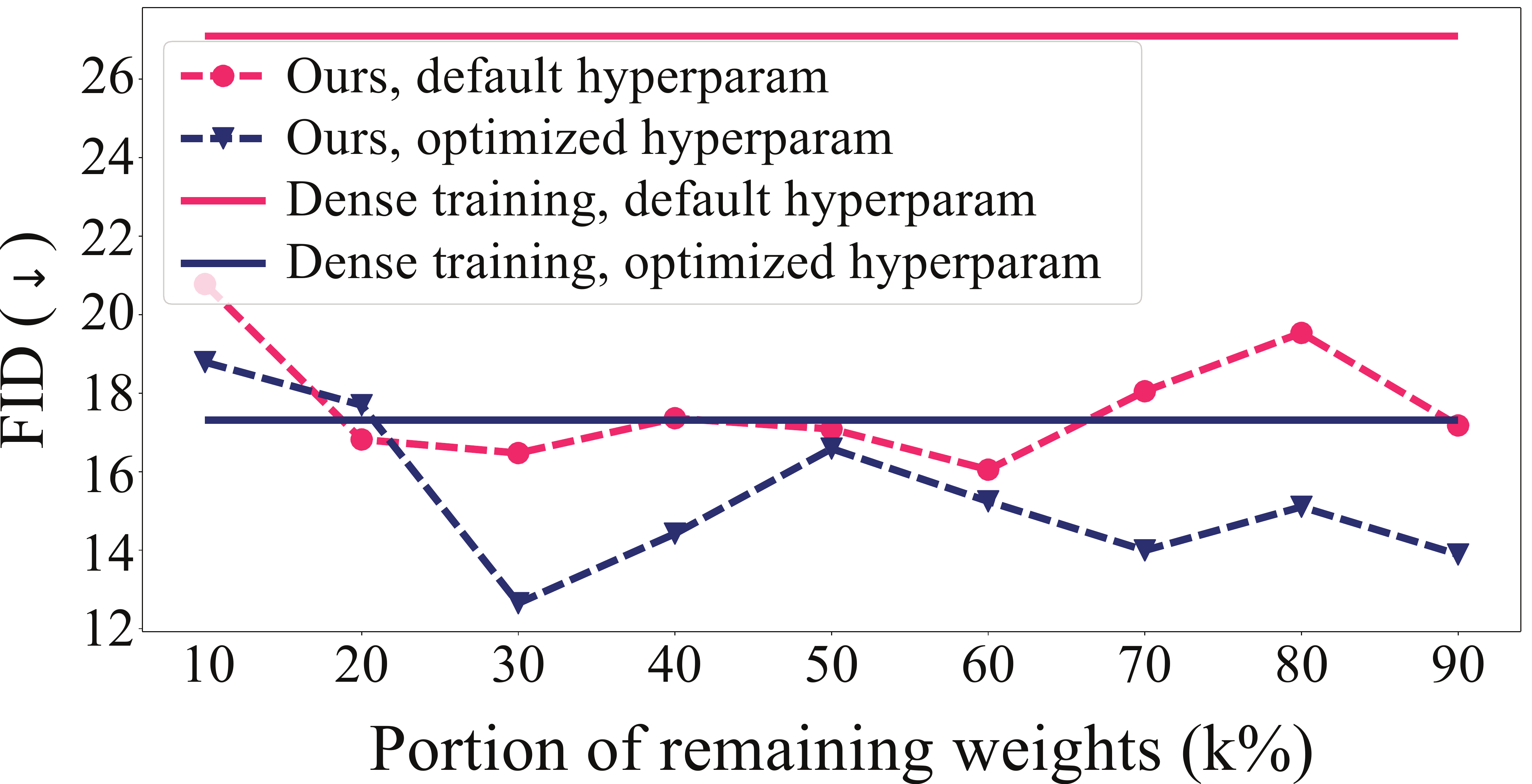}
\caption{
\textbf{The impact of applying our method on pretrained models (GFMN; LSUN-Bedroom).} Here, we test two pretrained models: (1) one using the default hyperparameter (solid red line)~\citep{santos2019learning}; (2) the other using a more optimized hyperparameter found from our experiments (solid blue line). Each dashed line shows the performance of the subnetwork obtained by our pruning method for each pretrained model with different target $k$. 
Our pruning method finds a subnetwork that performs better than the corresponding pretrained dense model in various sparsity.}
\label{fig:pretrained}
\end{figure}

\subsection{Experiment 2: Can we use our method to lighten the pretrained models?}

Recall that one of the biggest questions in the literature on model compression is whether we can find a good subnetwork within a fully trained model without losing performance~\citep{lecun1989optimal,HanEtAl16,wiedemann2020deepcabac,isik2021rate}. Focusing on this fundamental question that has been discussed over decades, we investigate whether our pruning method can provide a positive answer for this question in generative models.

Figure~\ref{fig:pretrained} shows the performance of our pruning method when applied to dense GFMN models trained on the LSUN dataset. Two pretrained models are considered in this experiment: (1) the model trained with the default hyperparameter suggested by~\citep{santos2019learning}, and (2) the model trained with the optimized hyperparameter found in our experiments. In both dense pretrained models, applying our pruning method maintains or even improves the performance when the portion of remaining weights is chosen within 10\% -- 90\% regime. 
This shows the practical importance of our method in that an off-the-shelf generative model (that already has reasonable performance) can be lightened to achieve 10x efficiency while having similar or better performance compared with the dense pretrained model.  

A natural follow-up question is whether having a better performance at the dense model implies having a better performance after applying our pruning method. 
The examples in Figure~\ref{fig:pretrained} show this is true in our experimental setting. However, exploring the answer to this question in various settings is out of the scope of this paper. We leave this as future work.

\subsection{Experiment 3: What happens when we further train a strong lottery ticket?}

The pioneering work~\citep{ramanujan2020s} on finding strong lottery tickets (SLTs) on discriminative networks had an interesting observation: SLTs found in their work are not fine-tunable, \textit{i.e.}, the performance of SLT does not improve even after weight training. 
We test whether this observation is true in SLTs found in generative models by fine-tuning the subnetwork found by our algorithm.

\begin{figure}[t]
\centering
\includegraphics[width=1\columnwidth ]{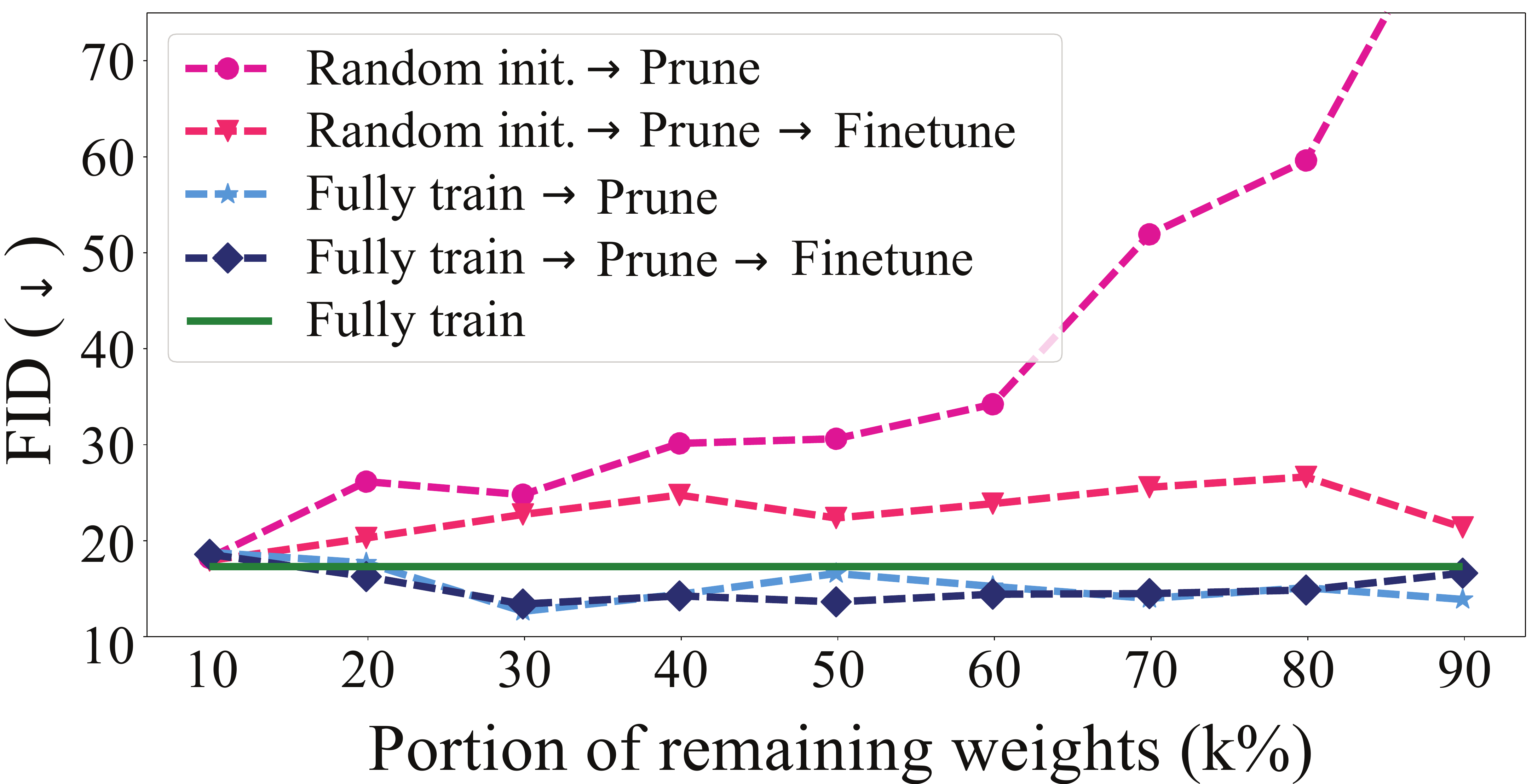}
\caption{
\textbf{The impact of fine-tuning on subnetworks (GFMN; LSUN-Bedroom).} We show the FID performances for two scenarios: (1) when the dense model is randomly initialized; (2) when the dense model is fully trained. 
}
\label{fig:retraining}
\end{figure}

Figure~\ref{fig:retraining} compares performances of the subnetwork (obtained by our pruning method) before and after the fine-tuning. We test on two subnetworks: one obtained from a randomly initialized network, and the other one obtained from the network having fully trained weights.  
For the subnetwork obtained from a randomly initialized network, fine-tuning the network improves the performance in all sparsity regimes except when the portion of remaining weights is $10\%$. The performance of the subnetworks obtained from a fully trained network does not improve even after fine-tuning in all sparsity regimes. From these experiments, one can confirm that subnetworks that already achieve the performance of the fully trained dense model cannot be improved by fine-tuning the survived weights. This coincides with the observation made in~\citep{ramanujan2020s}.

\section{Discussions}
Here, we provide some interesting discussion topics on finding strong lottery tickets in generative models.

\begin{figure}[t]
\centering
\includegraphics[width=1\columnwidth ]{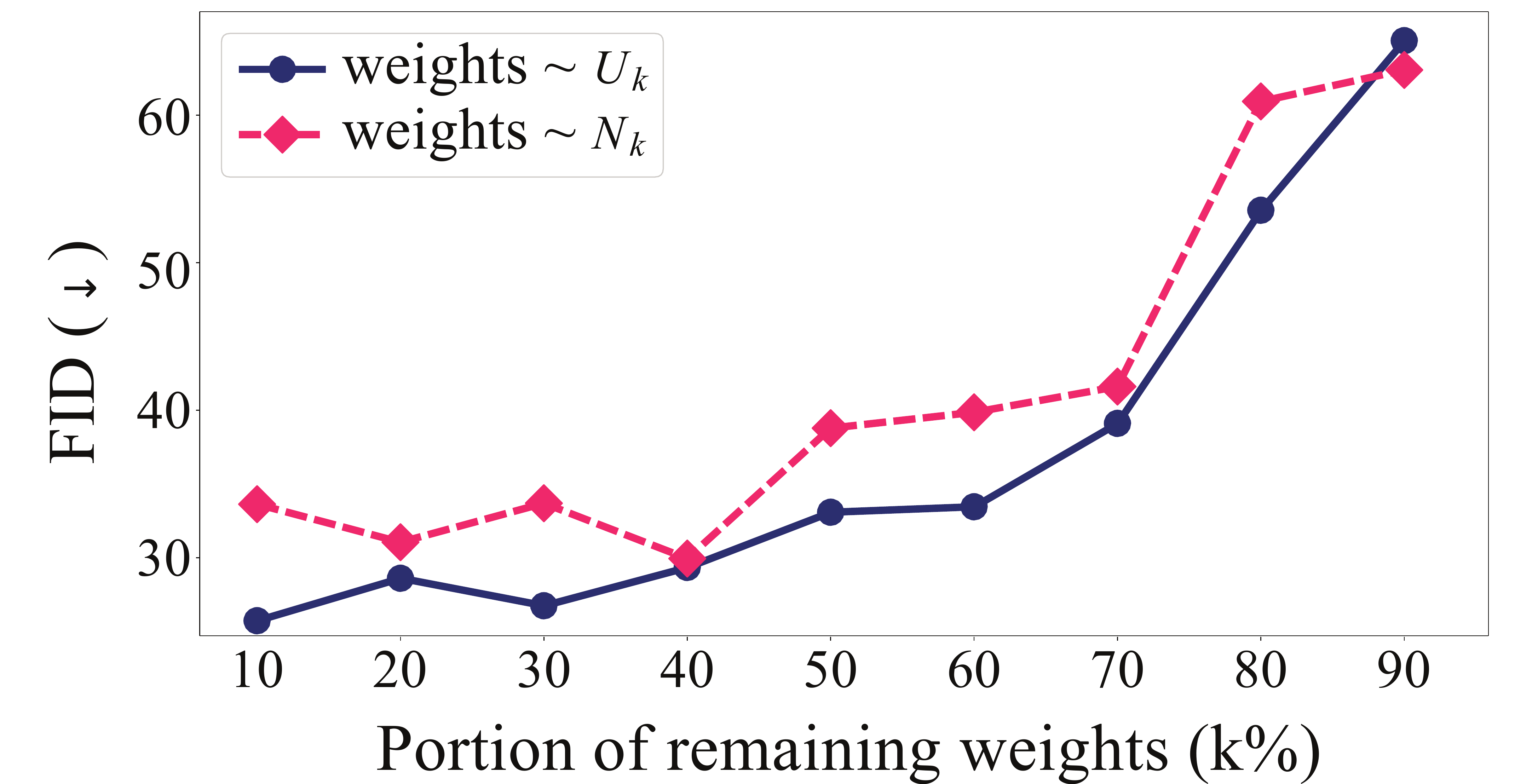}
\caption{ \textbf{The impact of various initializations (GFMN; LSUN-Bedroom).} We visualize the performance of our method according to different weight initializations: Kaiming normal $N_k$ and signed Kaiming constant $U_k$. The latter option shows better performance overall. }
\label{fig:varyingdis}
\end{figure}

\begin{table}[t]
\centering
\fontsize{9}{10}
\tabcolsep=0.1cm
\begin{tabular}{c|c|c|c|c|c|c}
\toprule
\begin{tabular}[c]{@{}c@{}}Ch. 
Multiplier ($n$)\end{tabular} & 0.4   & 1.0     & 1.4   & 1.6  & 1.8  & 2.0  \\
\midrule
FID ($\downarrow$)                                                            & 26.96 & 10.93 & 10.12 & 8.83 & 8.97 & 8.31 \\ 
\bottomrule
\end{tabular}
\caption{The impact of channel width on the performance of subnetworks (GFMN; CelebA).
The performance improves as the channel width increases.}
\label{table:varyingm}
\end{table}
\paragraph{Factor analysis}

The generative performance of the subnetwork found by our method depends on multiple factors.
The first factor is how we initialize weights of a neural network.
Figure~\ref{fig:varyingdis} compares two random weight initialization methods: ``Kaiming normal ($N_k$)" and ``signed Kaiming constant ($U_k$)". One can confirm that in most sparsity regimes, the signed Kaiming constant has better performance (lower FID value). This observation is consistent with the results in~\citep{ramanujan2020s} for discriminative models. Intuitively, weight initialization can be considered important because subnetwork search space varies depending on weight initialization. However, understanding which weight initialization can perform well still requires deeper research on the structure of neural networks.

The second factor is the channel width $n$ of the network. The default network is denoted by $n=1$, and we test on various networks having $n$ times larger channel width at each layer.
Table~\ref{table:varyingm} shows the performance of a strong lottery ticket found by our method for various $n$. 
One can confirm that as channel width increases, the subnetwork's performance improves. This observation makes sense: as the width increases, the number of subnetworks in a randomly initialized network increases exponentially; the probability of finding a better subnetwork therefore increases.

\begin{figure}[t]
\centering
\includegraphics[width=1\columnwidth ]{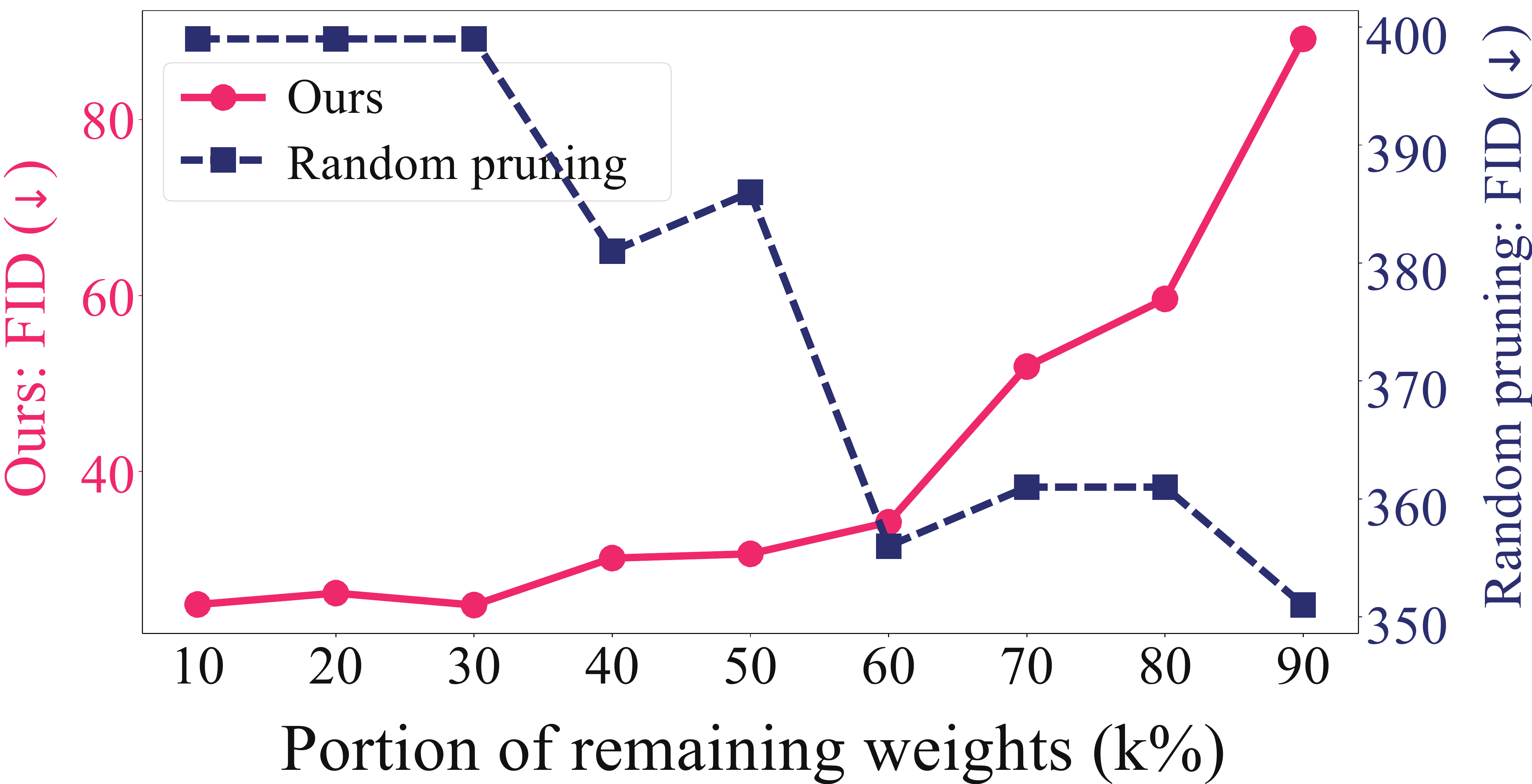}
\caption{\textbf{Performance comparison of random pruning versus our pruning method (GFMN; LSUN-Bedroom).} The performance gap in FID between ours and random pruning is huge. Unlike our method, random pruning fails to find SLT and loses performance as the portion of remaining weights gets decreased.}
\label{fig:random}
\end{figure}

\paragraph{Does it have to be the EP algorithm and the moment-matching score?} 
Because our method is the first algorithm to address prune-at-initialization in generative models, there is no base method for comparing. A possible na{\"i}ve alternative to our method would be random pruning; some might wonder if it has to be the EP algorithm and if it is also possible with random pruning. To address this point, we compare random pruning and our method and show that our results are not something that can be achieved by mere chance (see Figure~\ref{fig:random}). While our method finds a sufficiently meaningful subnetwork that produces realistic images, random pruning does not find such a subnetwork.

\begin{figure}[t]
\centering
\includegraphics[width=0.8\columnwidth ]{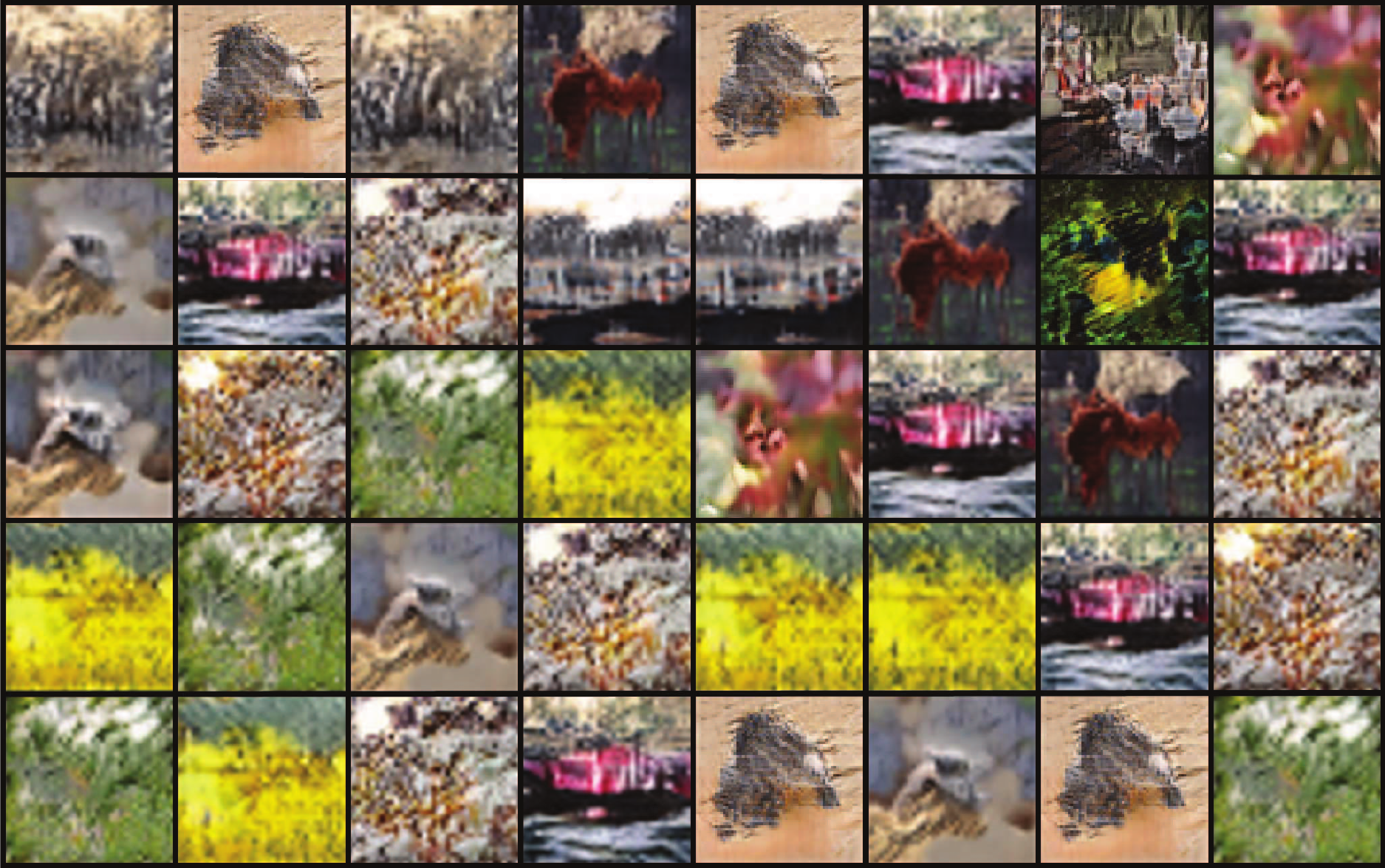}
\caption{ \textbf{Visualization of generated images with the adversarial loss (BigGAN, BabyImageNet).} Here, we leverage the adversarial loss to obtain scores for selecting a subnetwork mask. The resultant generator fails to generate diverse images and gets mode-collapsed.
}
\label{fig:advloss}
\end{figure}

Another natural follow-up question would be whether we can substitute the MMD score function with the adversarial loss, which is commonly used for training modern generative models. 
In our experiments, we find that when using the adversarial loss, network pruning simply fails due to its notorious instability as the score keeps changing (see Figure~\ref{fig:advloss}). 
Here, the subnetwork that we obtained via adversarial loss suffers from unrealistic image generation and mode-collapse phenomena.

\begin{figure*}[t]
\centering
\includegraphics[width=1.04\linewidth]{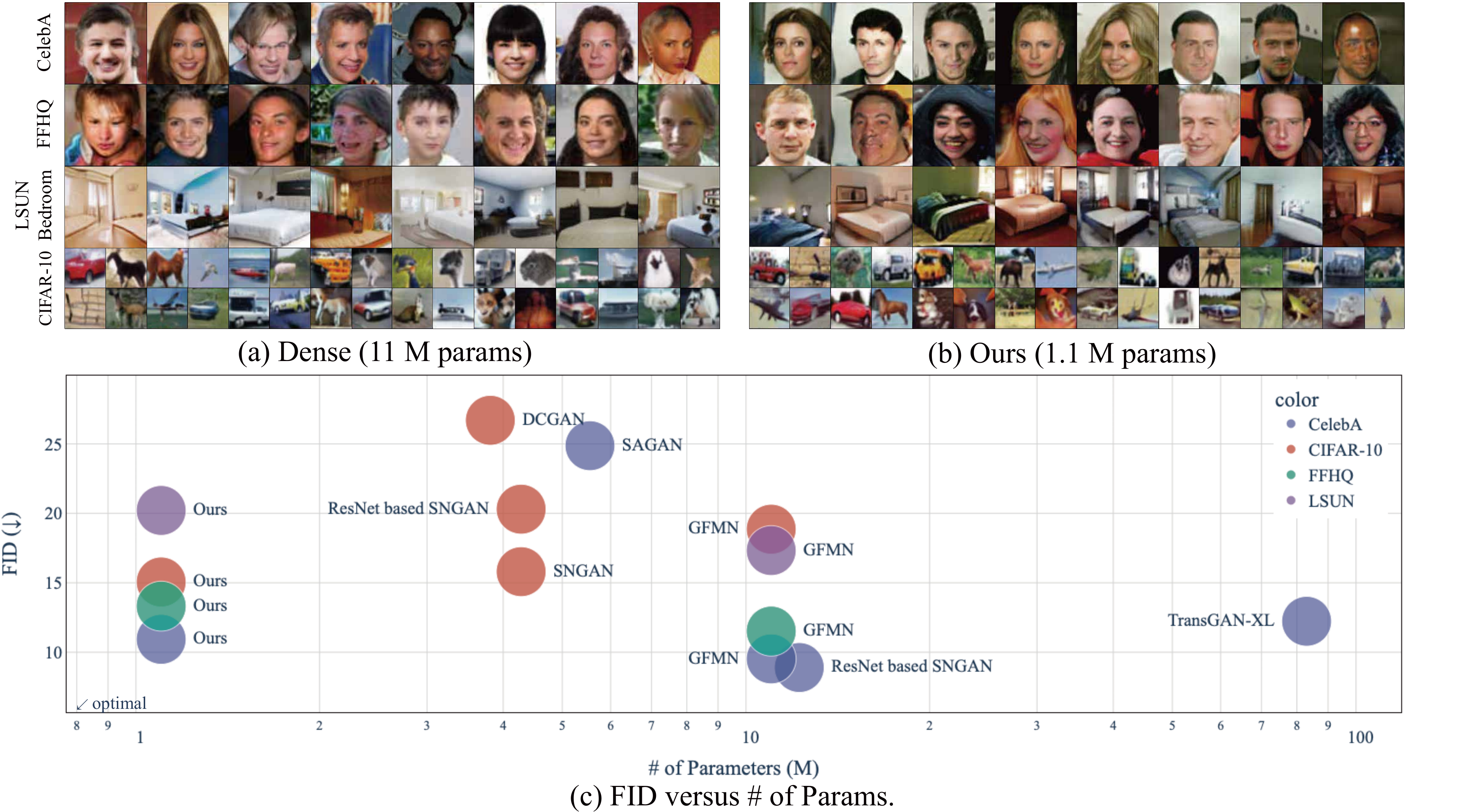}
\begin{tabular}{cc}
\end{tabular}
\caption{\textbf{Qualitative and quantitative comparison (GFMN; CelebA, FFHQ, and CIFAR-10).} Qualitative results for trained dense model (top left) and ours (top right). Ours can find SLT of good image quality and diversity. We also show that the GFMN generators pruned via ours show decent performance with a very small number of parameters (bottom). Note that the left bottom is the optimal point, where both the FID value and the number of parameter are small.
}
\label{fig:qualitative}
\end{figure*}

\paragraph{Efficient multi-domain generation} 
One nice property of our method is that it enables efficient multi-domain generation. Although several studies have proposed a model for multi-domain generation, all require a specific architecture to do so~\citep{2019FUNIT,2019StarGANv2, baek2021rethinking}. On the other hand, since our method only finds a subnetwork from the randomly initialized generator, one can use the same architecture and perform multi-domain generation simply by changing the mask found a priori. Unlike the other methods, our framework does not require any modification in the architecture to add a new domain to generate; one can simply find another mask, which is more efficient than developing a new model or fine-tuning the model architecture to a new domain. All results in Figure~\ref{fig:qualitative} on various domains (FFHQ, LSUN, CIFAR-10, and CelebA) are generated by simply changing the mask to the generator with the same weights. 
In Figure~\ref{fig:qualitative} (c), we show that the GFMN generators pruned via our method show decent performance in various datasets with a very small number of parameters.

\paragraph{Can we improve the performance of our method?}
In this paper, we shed light on the potential of finding strong lottery tickets in generative models by using the edge-popup (EP) algorithm and the MMD loss. Here, we discuss possible ways of improving the performance of our method.
Note that a recent work~\citep{sreenivasan2022rare} on pruning discriminative networks found that there are two methods to improve the performance of EP: (1) using global EP (pruning weights by sorting the scores globally) instead of vanilla EP (pruning weights by sorting the scores at each layer), and (2) using gradual pruning (moving from dense regime to sparse regime gradually during pruning) instead of vanilla EP which moves to the sparse regime from the beginning.
Inspired by this observation, we expect applying EP with these two variants (global pruning and gradual pruning) in our method has the potential to improve the performance of the SLT in generative models.

\paragraph{Are there strong lottery tickets in high resolution?}
To explore SLT at high resolution, we apply our method for 128 resolution and observe that the trend remains the same. That is, we can also find SLT in high-resolution generative models. We include this result in Appendix E. %Unfortunately, 
Although we can find SLT at high resolution, we observe that our method converges slowly due to high-dimensional feature maps. Consequently, the performance of the dense network is not satisfactory. We can improve performance by investigating stronger kernels, but this is beyond the scope of this paper and is an interesting future research direction.

\paragraph{Why is pruning generator challenging?}
While pruning discriminative models has shown remarkable results, pruning generators still has various challenges: 1) There are no obvious criteria. Unlike supervised learning which has labels, it is difficult to provide clear criteria for what to prune in generative models. 2) Training is unstable. Many studies have shown that pruning generators severely reduces performance due to training instability. 3) Generative models are mostly decoder structures. Pruning the weights in the decoder can have a more significant impact on the final output than the encoder due to the expanded output space. In this paper, our framework provides obvious criteria and obtains pruned generators stably. Appendix D contains the results showing the stability of our method.

\section{Related Work}
\paragraph{Neural Network Pruning.}
Conventional iterative train-prune-retrain framework incurs massive \emph{training costs}, even though it significantly reduces computational costs at test time. To address this issue, two categories of network pruning methods have been suggested in recent years:
(1) pruning a random network in a way that the subnetwork is trainable to have a good performance, and (2) pruning a random network in a way that the subnetwork itself is having good performance without any weight updates. A subnetwork obtained from the former category is called \emph{weak} lottery ticket~\citep{frankle2018lottery} and a subnetwork in the latter category is called \emph{strong} lottery ticket~\citep{ramanujan2020s}. There have been extensive works on developing theories and algorithms on weak/strong lottery tickets in discriminative networks~\citep{frankle2018lottery,frankle2020linear,lee2018snip,wang2019picking,tanaka2020pruning,frankle2020pruning,su2020sanity,ma2021sanity,chen2020lottery}, but results on generative models were limited so far.

\paragraph{Compressing generative models.}
Some recent works focus on finding weak lottery tickets for obtaining lightweight generative models. 
For GANs and VAEs, the authors of~\citep{kalibhat2021winning,chen2021gans} used iterative magnitude pruning~\citep{frankle2018lottery} a popular unstructured pruning method.
However, unlike \emph{strong} lottery tickets which do not need any weight updates, \emph{weak} lottery tickets require additional weight updates for achieving reasonable performance. Our work differs from these works in two perspectives: (1) we focus on finding strong lottery tickets that perform well without any weight update, and (2) we do not rely on GAN loss, thus finding a good subnetwork stably.
To the best of our knowledge, the present paper is the first work that shows the existence of a strong lottery ticket in generative models.

\section{Conclusion}
In this paper, we investigated strong lottery tickets (SLT) in generative models. 
While all existing works on building lightweight generative models suffer from huge weight update costs, performance degradation, limited generalizability, or complicated training, 
we circumvented these problems by searching for SLT. To the best of our knowledge, we are the first to show the existence of SLT in generative models; SLT was previously observed only in discriminative models. By exploiting the moment-matching approach for scoring important weights in a randomly initialized generator, our framework finds SLT stably without bells and whistles. 
Our experimental results showed that our method could successfully find a sparse subnetwork in various datasets, and the discovered subnetwork achieved similar or even better performance than the trained dense model even when only 10\% of the weights remained.

\section*{Acknowledgments}
This work was supported by the National Research Foundation of Korea (NRF) grant funded by the Korea government (MSIT) (No. 2.220574.01), Institute of Information \& communications Technology Planning \& Evaluation (IITP) grant funded by the Korea government (MSIT) (No.2020-0-01336, Artificial Intelligence Graduate School Program (UNIST)), and Institue of Information \& communications Technology Planning \& Evaluation (IITP) grant funded by the Korea government (MSIT) (No.2022-0-00959, (Part 2) Few-Shot Learning of Causal Inference in Vision and Language for Decision Making).

\bibliography{reference}

\onecolumn
\begin{center}
{\LARGE\bfseries Can We Find Strong Lottery Tickets in Generative Models?\\- Supplementary Materials -}
\end{center}
In this document, we describe our network architecture, implementation details, and additional experimental results on other models for our paper, ``Can we Find Strong Lottery Tickets in Generative Models". 
\\
\section{Appendix A. Experimental Results on Other Models}

\paragraph{Pruning randomly initialized networks.}

\begin{figure*}[ht!]
\centering
\begin{tabular}{cc}
\includegraphics[width=0.45\linewidth]{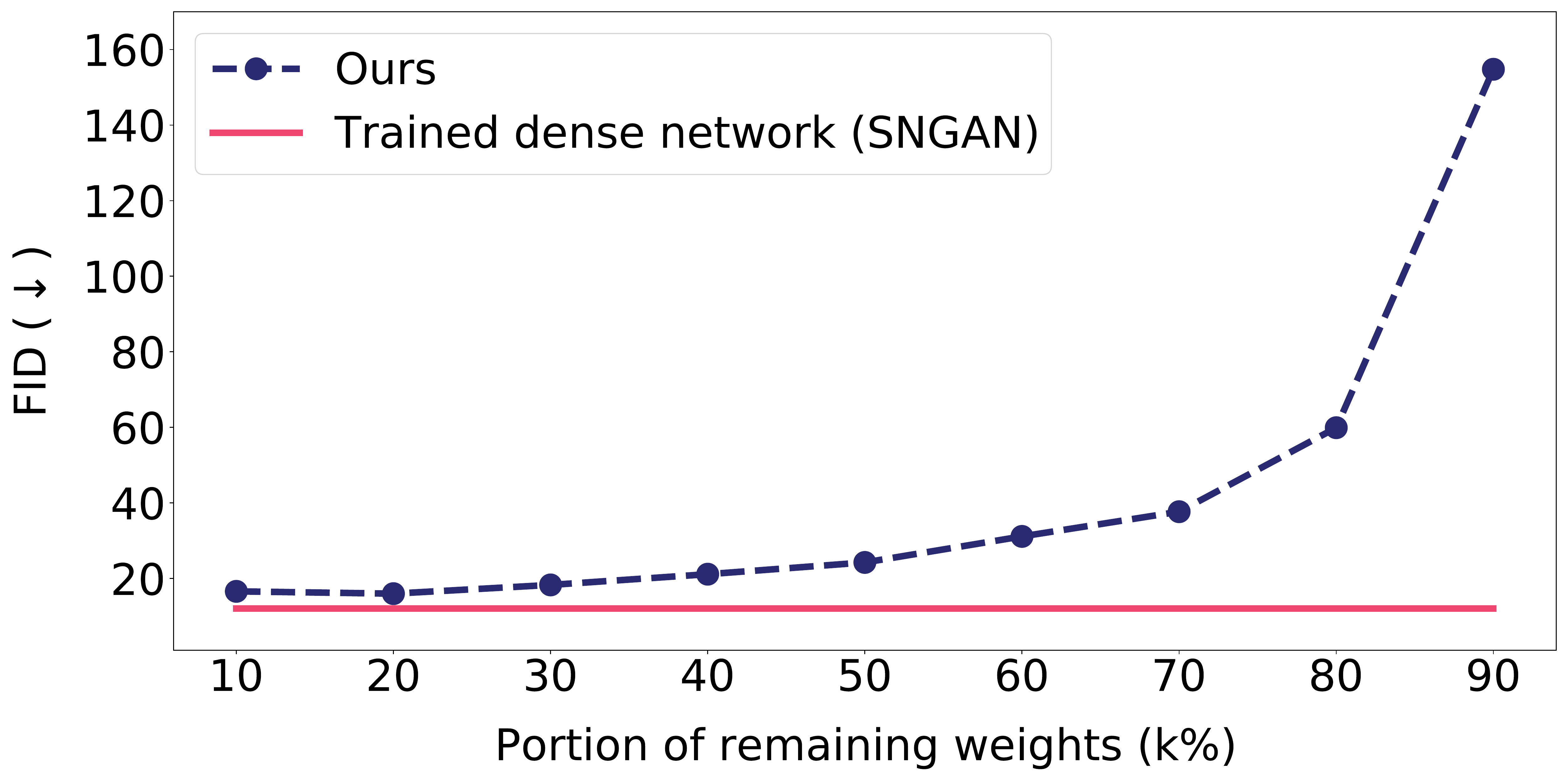}  &
\includegraphics[width=0.45\linewidth]{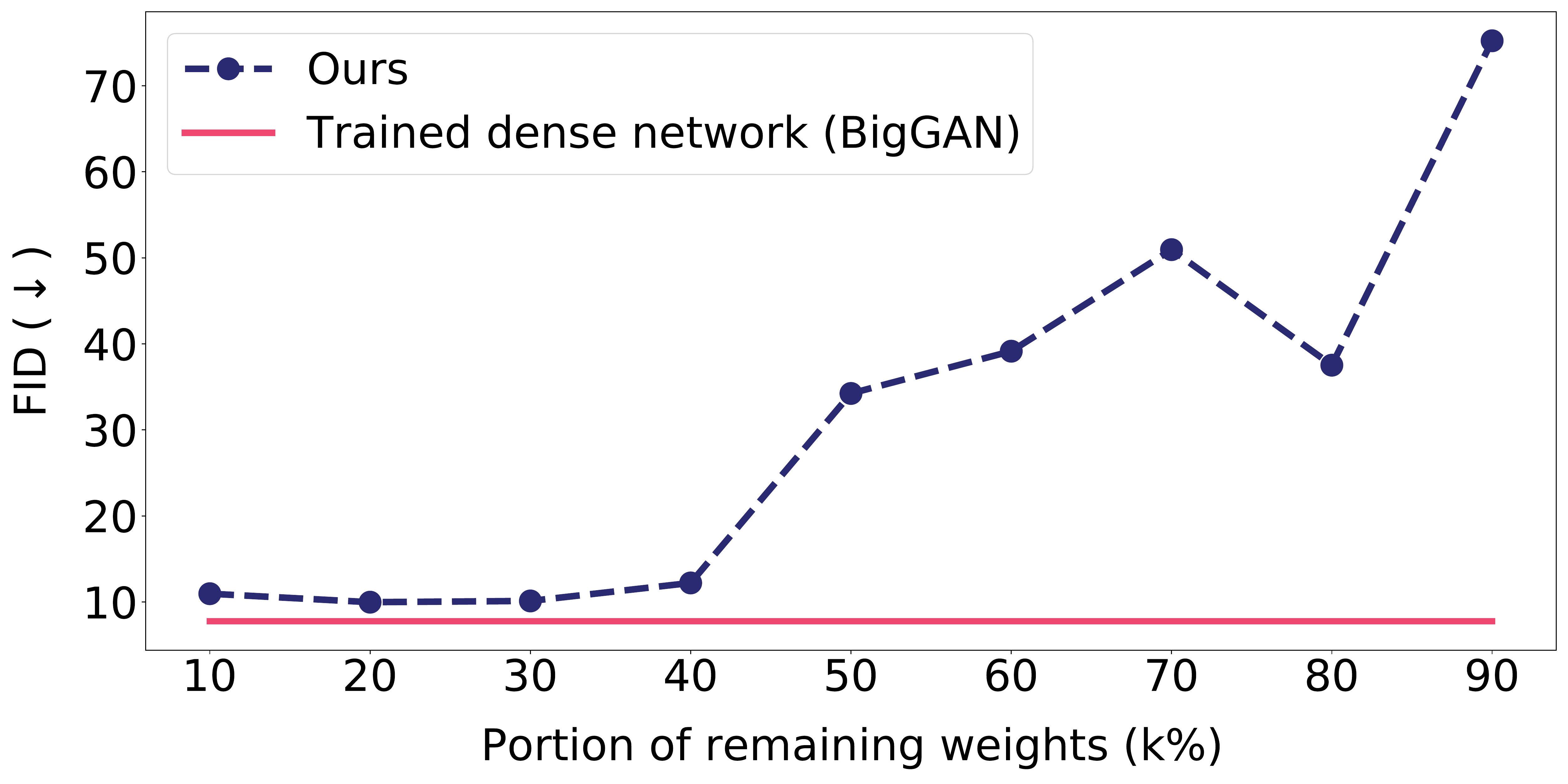}
\end{tabular}
\caption{ {\bf Comparison of FID of (1) a pretrained dense model and (2) the subnetworks of a randomly initialized network using our pruning method (left: SNGAN, right: BigGAN; CelebA).}
The performance of our method improves as $k$ decreases. This shows that our method successfully finds good subnetworks from a randomly initialized model. We use dense models of SNGAN with a FID score of 12.078 and BigGAN with a FID score of 7.756, respectively.
}
\vspace{-0.4 cm}
\label{fig:random2}
\end{figure*}

In Figure 3 in the main manuscript, we compared the FID values of (1) trained dense model and (2) randomly initialized subnetworks obtained by our pruning method for the LSUN-bedroom dataset. We perform the same experiments on SNGAN and BigGAN on CelebA in Figure~\ref{fig:random2}.
Similar to Figure 3 in the main manuscript, the FID value of subnetworks obtained by our method decreases as $k$, and the portion of remaining weights decreases from 90\% to 10\%. When most random weights remain (\textit{e.g.}, $k=90\%$), the pruned network is equivalent to the \textit{randomly initialized} model. So it is natural that the performance is bad in now sparse regime. On the other hand, When most random weights are pruned (\textit{e.g.}, $k=10\%$), we can find SLT, which has fine generative performance without weights training. 

\paragraph{Pruning pretrained networks.}

\begin{figure*}[ht!]
\centering
\begin{tabular}{cc}
\includegraphics[width=0.45\linewidth]{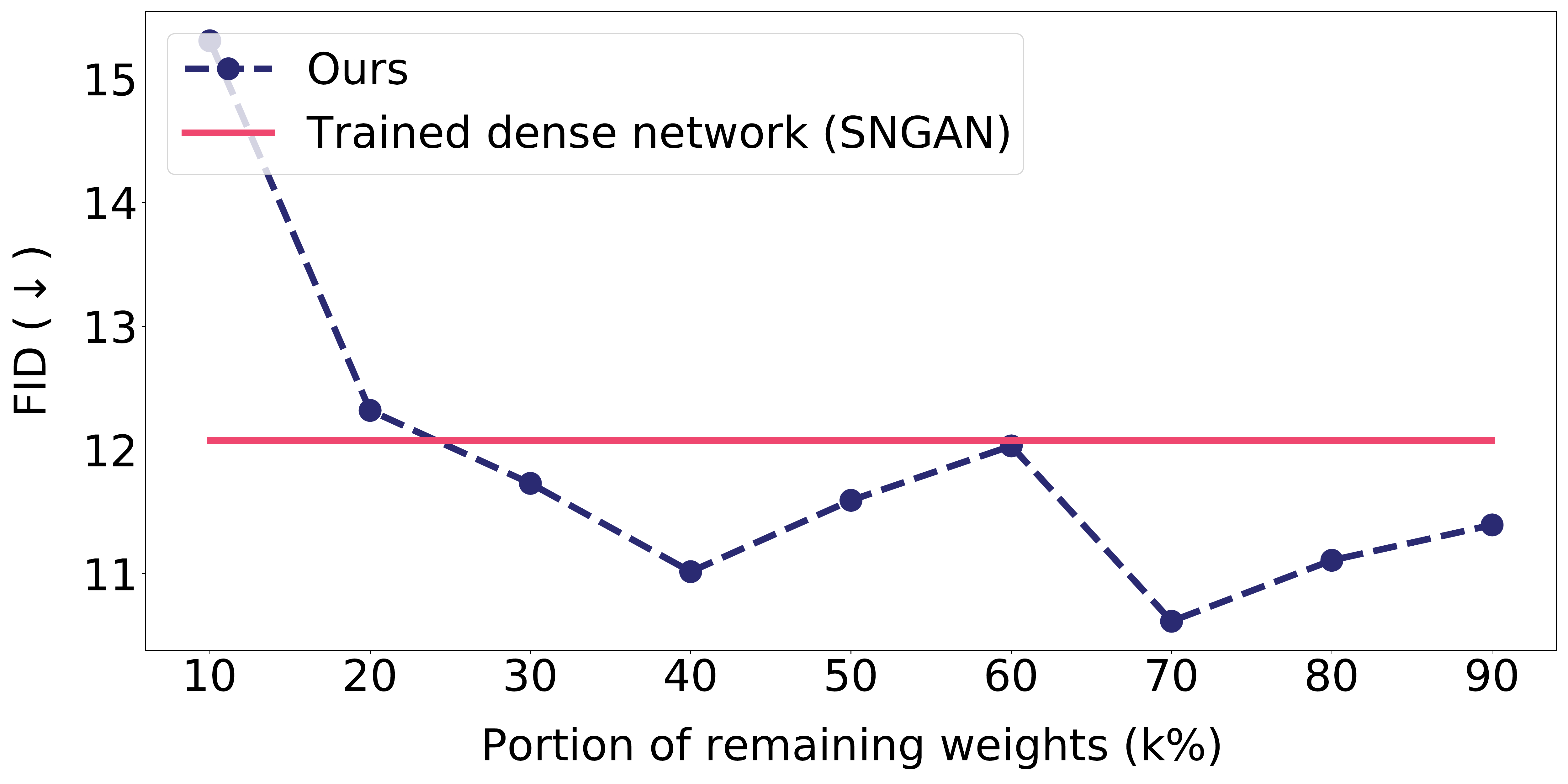}  &
\includegraphics[width=0.45\linewidth]{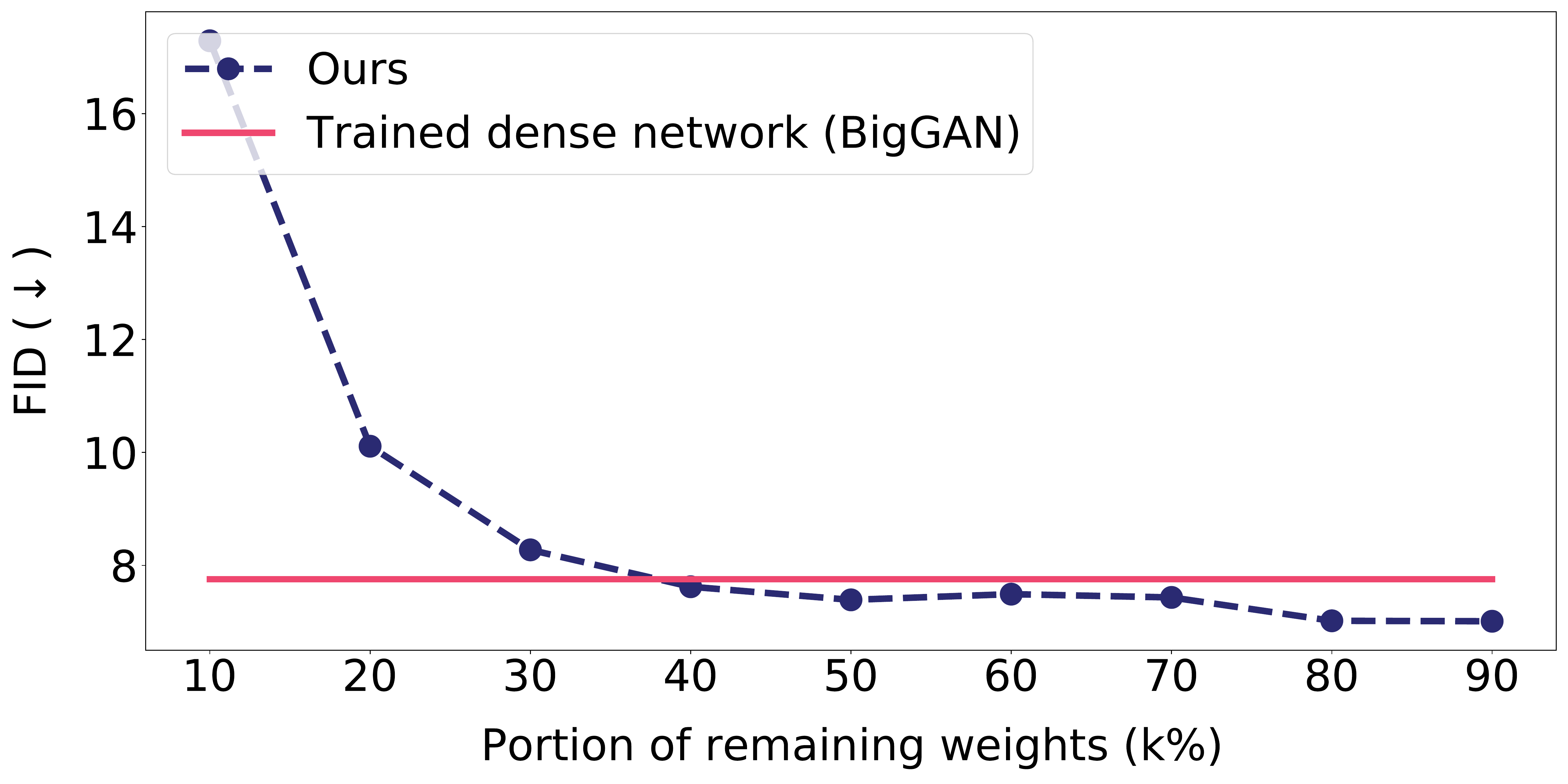}
\end{tabular}
\caption{ {\bf Comparison of FID of (1) a pretrained dense model and (2) the subnetworks obtained by pruning the pretrained dense model in (1) using our method (left: SNGAN, right: BigGAN; CelebA).}
Our method finds subnetworks that perform better than pretrained dense models, respectively, when $k$ is larger than 20$\%$ on SNGAN and when $k$ is larger than 30$\%$ on BigGAN. Notably, we identify that subnetworks, which outperform or match performance of dense networks by remaining about 30$\%$ of the weights on SNGAN and 40$\%$ of the weights on BigGAN, respectively. This shows our method initially designed for pruning a randomly initialized model can be applied to improving the performance of a pretrained model as well. We use dense models of SNGAN with a FID socre of 12.078 and BigGAN with a FID score of 7.756, respectively.
}
\vspace{-0.4 cm}
\label{fig:pretrain}
\end{figure*}

Figure 5 in the main manuscript showed FID values of (1) the trained dense model and (2) the subnetwork of (1) obtained by our pruning method for the LSUN-bedroom dataset. We perform the same experiments on SNGAN and BigGAN on CelebA in Figure~\ref{fig:pretrain}.
Similar to Figure 5 in the main manuscript, the FID value of subnetworks obtained by our method is lower than that of the trained dense model for $k$ larger than 30\%. Unlike randomly initialized networks, pruned networks show better performance with more weights remaining. In contrast to the previous case, when most weights remain (\textit{e.g.}, $k=90\%$), the pruned network is equivalent to the \textit{pretrained} model. We can find subnetworks, which outperform or match performance of dense networks by applying our method to pretrained networks.

\section{Appendix B. Experimental Setup}

We provide detailed experimental settings including the generator architecture and training hyperparameters.

\paragraph{Description of ResNet generator architecture.}
The architecture of our generator and detailed setup follows \citep{santos2019learning}.
ResNet generator consists of four residual blocks with ReLU, BatchNorm, Upsample, and Conv. Here, we use the nearest upsampling module with the scale factor 2. Conv (convolution layer) in the residual block has $3\times3$ kernel size, $\textit{stride}=1$ and $\textit{padding}=1$, and the convolution for residual connection has $1\times1$ kernel size, $\textit{stride}=1$ and $\textit{padding}=0$.

\paragraph{GFMN and Edge Popup setting.}

We use the ImageNet-pretrained VGG19 \citep{simonyan2014very} released in PyTorch Torchivision\footnote{\url{https://pytorch.org/vision/stable/models.html/}} and extract the features from the first 16 ReLU layers.

For pruning or training dense models, we use the Adam optimizer~\citep{kingma2014adam} with $\beta_1=0.5$, $\beta_2=0.999$ for weights or scores training following~\citep{santos2019learning}. The learning rate is set to $\text{lr}=0.00005$ scheduled with the cosine annealing method~\citep{loshchilov2016sgdr}. Training is performed for 2,000,000 iterations, and we use a batch size of $64$ in all experiments. To train stably and preserve better image quality in a small batch size regime, Adam moving average is used to calculate the moment of the real images with $\beta_1=0.00001$, $\beta_2=0.999$, which is the default setup of \citet{santos2019learning}. To show the existence of SLT, we freeze all the parameters (including Batch normalization parameters) of the generator.
When we apply our pruning method to randomly initialized model, we initialize the weights of generator using two distributions: ``Kaiming normal ($N_k = N(0, \sqrt{2/n_{l-1}}$)" and ``signed Kaiming constant ($U_k = \{-\sigma_k, \sigma_k\}$)" where pruning setting. When training dense models, the weight of generator is initialized with ``Xavier uniform ($U_x = \{-\sqrt{\frac{6}{n_{in}+n_{out}}}, \sqrt{\frac{6}{n_{in}+n_{out}}}\}$)".

\section{Appendix C. Evaluation Setting}

We compare randomly sampled 10K reference images and 10K generated images for evaluation of generative performance. We use the InceptionV3~\citep{szegedy2016rethinking} models for extracting features of images to evaluate them. The InceptionV3 model is provided by Pytorch official\footnote{\url{https://pytorch.org/hub/pytorch_vision_inception_v3/}}. We use Precision and Recall, Density and Coverage, and Fr\'echet Inception Distance (FID) to evaluate performance of network. We can observe that our experiments show consistent trend regardless of Precision \& Recall, Density \& Coverage, or FID during evaluation. The following evaluation metrics are what we use.

\paragraph{Fr\'echet Inception Distance.} Fr\'echet Inception Distance (FID)~\citep{heusel2017gans} computes the discrepancy of mean and covariance between the feature distributions of reference samples and fake samples. The distributions are assumed to be a multivariate normal distribution. The small FID score means that fake distribution approximates the real distribution well. 

\paragraph{Precision and Recall.} Precision and Recall~\citep{kynkaanniemi2019improved} evaluate quality and diversity separately for reference and fake samples. The high Precision means that generative models can generate high quality images and the high Recall means that generative models can generate diverse images. 

\paragraph{Density and Coverage.} Density and Coverage~\citep{naeem2020reliable} are improved metrics upon the Precision and Recall. They are robust for outliers, and their computational cost is lower than Precision and Recall. The high Density means that generative models can generate high quality images and the high Coverage means that generative models can generate diverse images.

\begin{figure*}[t!]
\centering
\includegraphics[width=0.4\paperwidth]{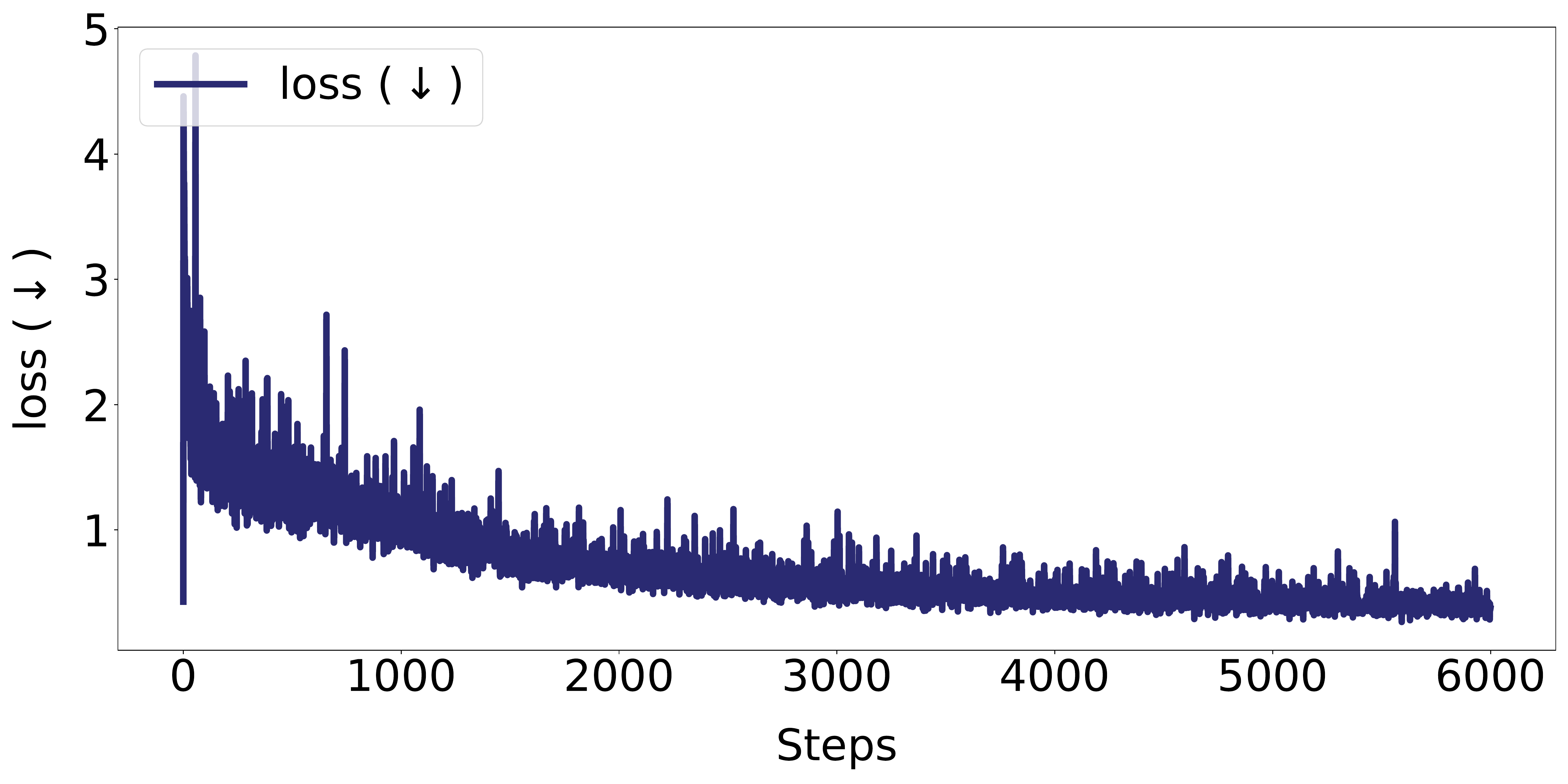}
\includegraphics[width=0.4\paperwidth]{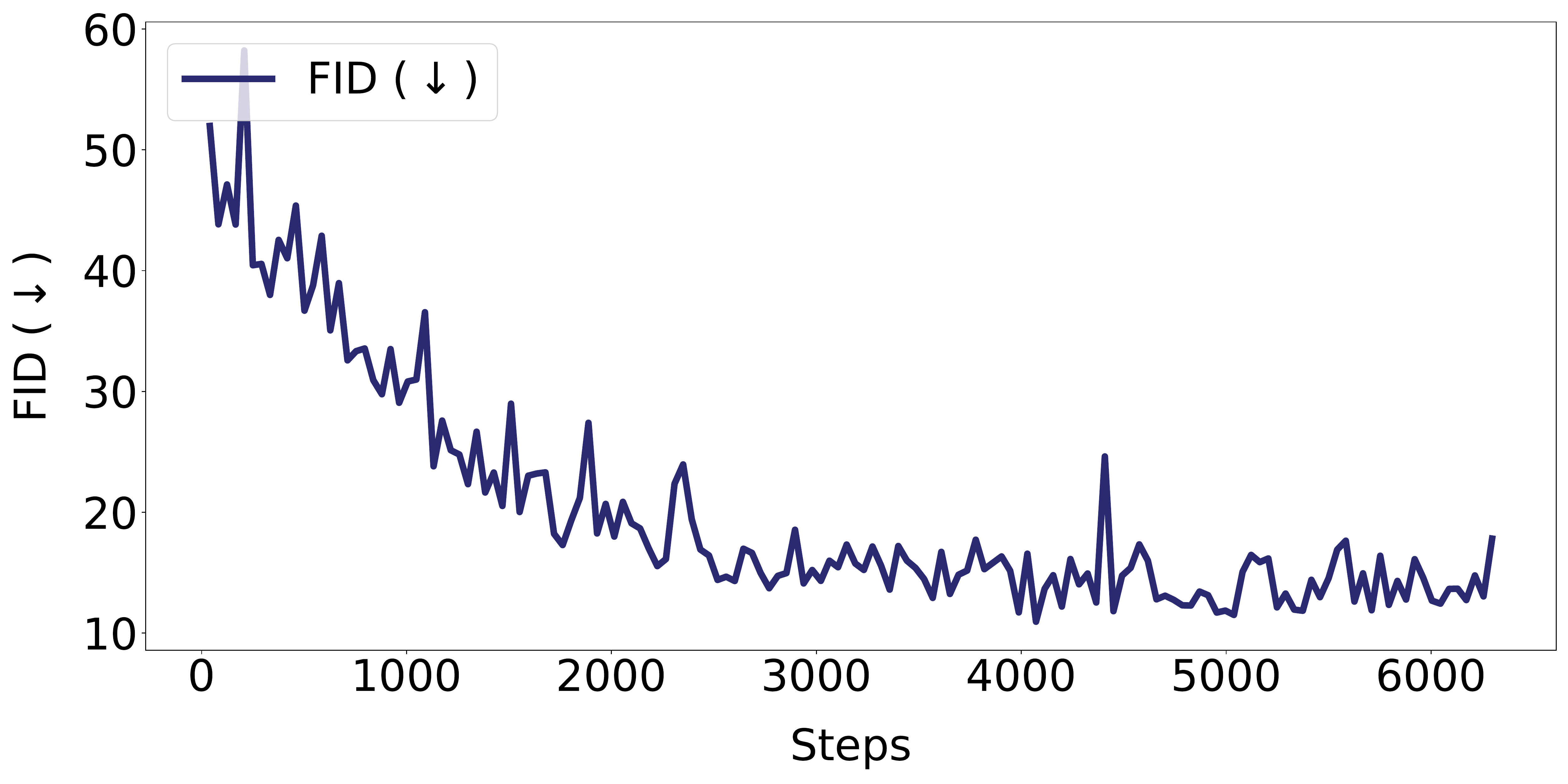}

\begin{tabular}{cc}
\end{tabular}
\caption{\textbf{Loss and FID convergence plot (GFMN; CelebA, remaining parameters = 10\%).} Our pruning method uses the MMD loss, thus, it converges stably and ensures the FID performance improvement as loss decreases.}
\label{fig:convergeplot}
\end{figure*}

\section{Appendix D. Additional Experimental Results}

Here we provide additional experimental results showing the advantages of our algorithm.

\begin{table*}[ht!]
    \centering
    \small
    \tabcolsep=0.33cm
	\begin{tabular}{c|ccccccc}
		\toprule
		Method & k = 10 $\%$ & k = 20 $\%$  & k = 30 $\%$  & k = 40 $\%$& k = 50 $\%$& k = 60 $\%$& k = 70 $\%$  \\
		\midrule

	   Edge Popup & 16.60 & 15.99 & 18.30 & 21.14 & 24.25 & 31.14 & 37.68 \\
         Global Edge Popup & 15.47 & 17.16 & 17.54 & 16.63 & 14.24 & 12.77 & 21.00    \\

		\bottomrule
	\end{tabular}%
    \caption{ {\bf Comparison of FID of vanilla Edge Popup and Global Edge Popup on randomly initialized SNGAN on CelebA.} Subnetworks obtained by global EP outperform subnetworks obtained by vanilla EP at all sparsity except k = 20$\%$.
    }
	\label{table:epvar_ran} 
\end{table*}

\begin{table*}[ht!]
    \centering
    \small
    \tabcolsep=0.33cm
	\begin{tabular}{c|cccc}
		\toprule
		Method & k = 10 $\%$ & k = 20 $\%$  & k = 30 $\%$  & k = 40 $\%$  \\
		\midrule
	   
	   Training (Dense) & 12.078 & 12.078 & 12.078 & 12.078\\
    \midrule
	   Edge Popup & \textcolor{blue}{15.309} & \textcolor{blue}{12.320} & \textcolor{red}{11.730} & 11.016 \\
         Global Edge Popup & \textcolor{red}{14.558} & \textcolor{red}{12.304} & 12.283 & \textcolor{blue}{10.673}    \\
         Edge Popup + Layer Freeze (First \& Last) & 34.926 & 15.005 & 12.224 & 10.833\\
         Global Edge Popup + Layer Freeze (First \& Last) & 39.702 & 14.433 & 12.102 & 11.022\\
         Global Edge Popup + Layer Freeze (Last) & 20.253 & 12.495 & \textcolor{blue}{11.759} & \textcolor{red}{10.655}\\
		\bottomrule
	\end{tabular}%
    \caption{ {\bf Comparison of FID of variations on fully trained dense SNGAN on CelebA dataset.} At extremely sparse region (\textit{e.g.}, $k=10\%$), Global EP without layer freeze shows best performance. On the other hand, Global EP with last layer freeze shows best performance at sparsity = 40$\%$. Red and blue indicate the best and the second best.}
	\label{table:epvar_pre} 
\end{table*}

\paragraph{Convergence plot}

We investigate how well our method converges in terms of the MMD loss and the FID. 
In Figure~\ref{fig:convergeplot}, we show the result when we prune a randomly initialized model to find a subnetwork with 10\% remaining parameters for the CelebA dataset. 
One can confirm that our method converges stably and that FID also decreases when MMD loss decreases.

\paragraph{Variations of Edge Popup} We also test two variants of EP; (1) global EP and (2) layer freeze EP. Recall that the vanilla EP prunes weights by sorting the scores at each layer. On the other hand, global EP prunes weights by sorting the scores globally across all layers. When we prune a fully trained network, particular layers may contain important information. For example, the first layer and the last layer are directly related to the latent and pixels, respectively. Therefore, we consider the layer-freeze EP defined as follows: for a portion of layers (\textit{e.g.}, first layer, last layer, or both) we do not apply pruning and use 100$\%$ of the trained weights, and for other layers we apply normal EP algorithm to prune weights.
In Table~\ref{table:epvar_ran}, we compare the performance of vanilla EP and global EP on randomly initialized SNGAN on CelebA. We can observe that global EP outperforms vanilla EP.
In Table~\ref{table:epvar_pre}, we compare the performance of variations of edge popup on fully trained dense SNGAN on CelebA. While layer freeze EP obtains poor subnetwork, global EP without layer freeze outperforms other methods at the sparsity of 10$\%$-20$\%$. In this result, we observe that subnetwork obtained by layer freeze EP has to have extremely sparse hidden layer to maintain the total sparsity when $k=10\%$ or $20\%$. Since all weights of freezed layers remain, sparsity ($k$) is close to 10$\%$ without the weight of the hidden layer. As $k$ increases, we can observe that performance of subnetwork obtained by layer freeze EP starts to be restored. Combination of global EP and last layer freeze outperforms other methods at the sparsity of 40$\%$.

\paragraph{Different feature extractors}
When estimating the data distribution by matching moments via MMD loss, its performance depends on how powerful the kernel is; our method will benefit from a better feature extractor. In GFMN~\citep{santos2019learning}, the authors tested various networks and reported that VGG shows the best result. Interestingly, we also found that VGG gives the best FID among VGG, ResNet, and ViT (9.4, 12.7, and 32 on CelebA; 28.3, 46.5, and 55.8 on LSUN Bedroom, respectively). Based on these results, we used VGG; but exploring the impact of the kernel in our method is an interesting future direction.

\paragraph{Qualitative results}

\begin{figure}[ht!]
\centering
\includegraphics[width=1\columnwidth ]{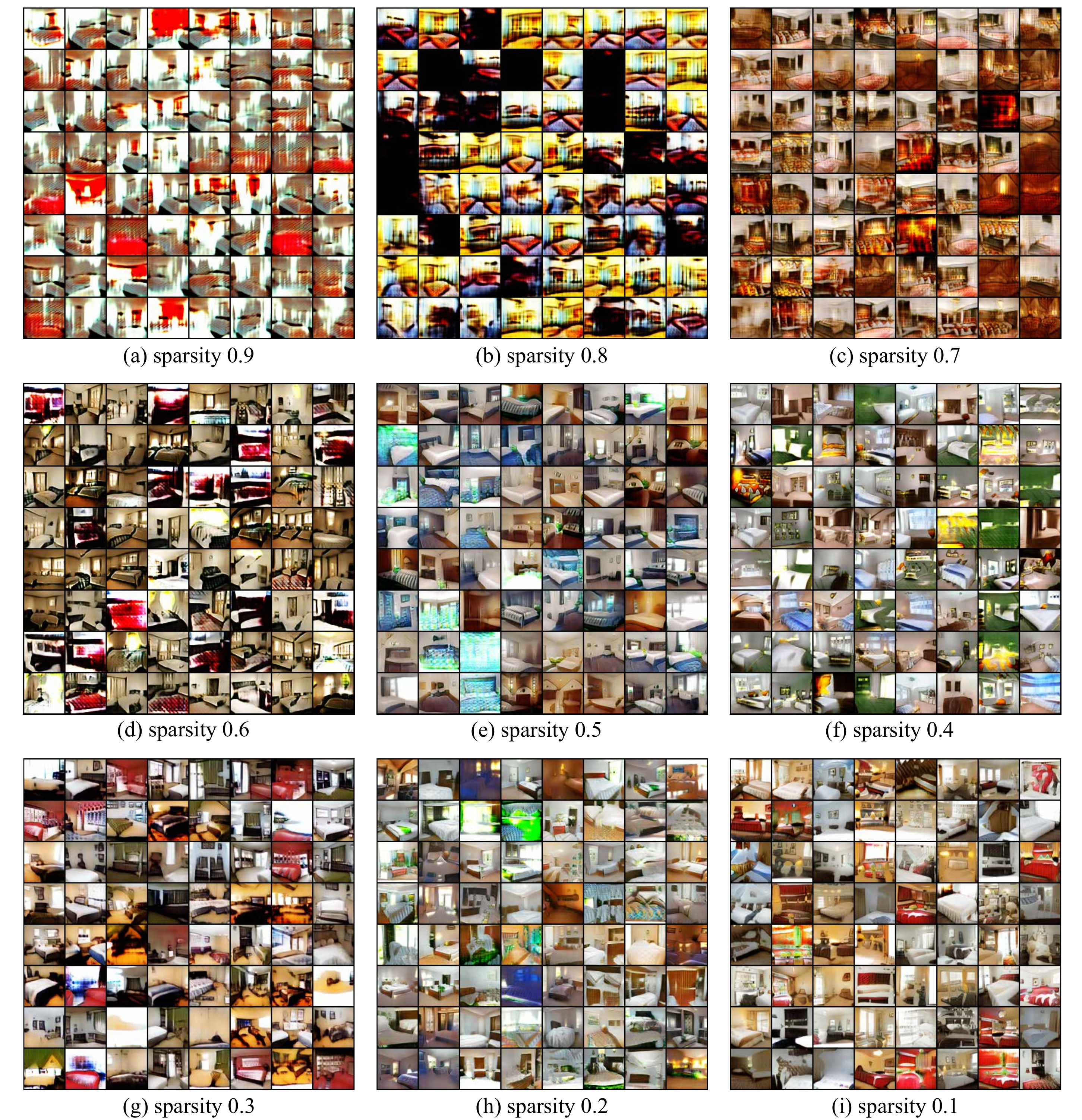}
\caption{ {\bf Qualitative results according to sparsity, pruning a random weight GFMN on the LSUN-Bedroom dataset.} We can observe that the pruned GFMN shows the best performance at sparsity = 10\%.
} 
\label{Random_LSUN_Bedroom}
\end{figure}

\begin{figure}[ht!]
\centering
\includegraphics[width=1\columnwidth ]{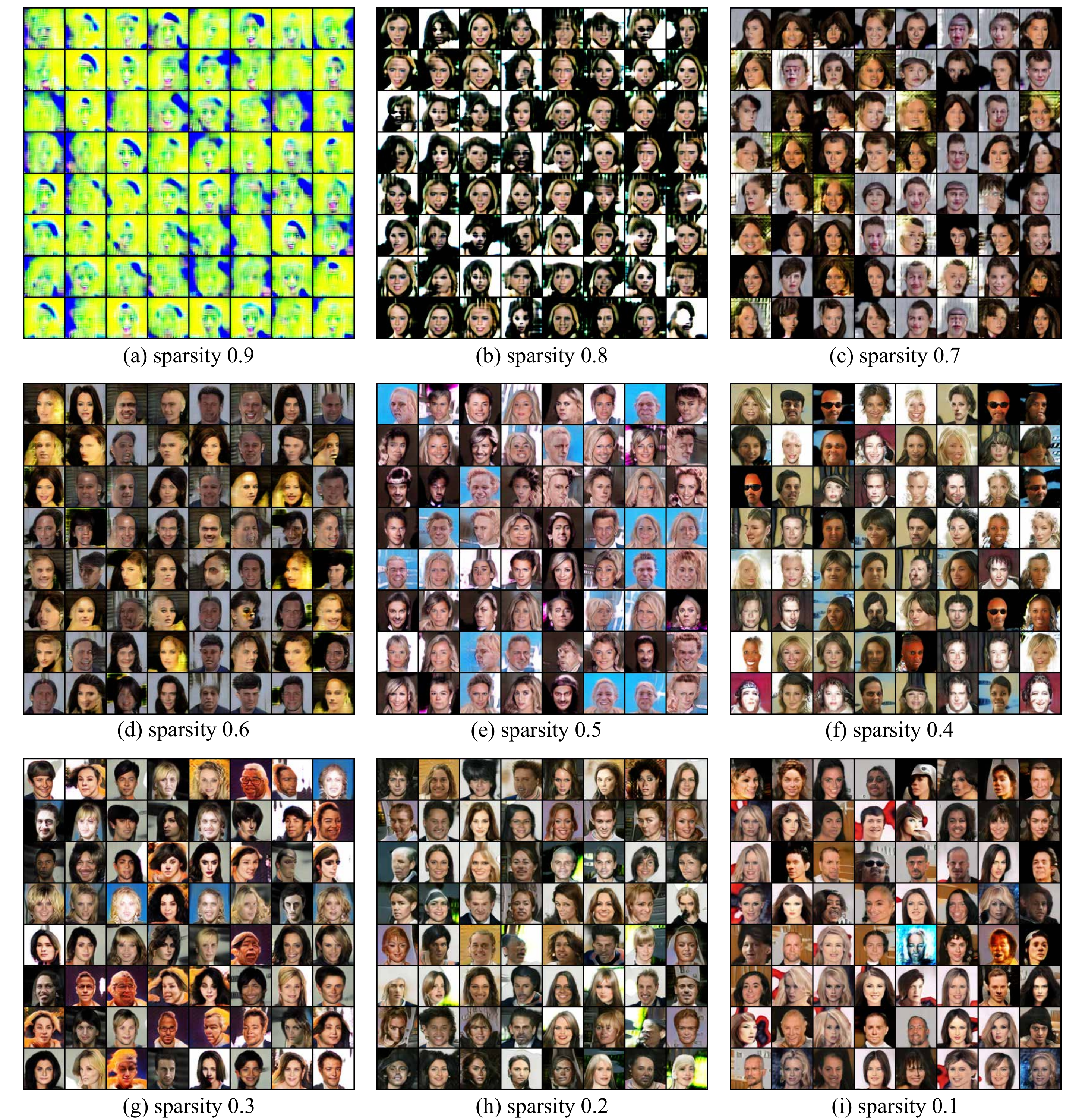}
\caption{ {\bf Qualitative results according to sparsity, pruning a random weight SNGAN on the CelebA dataset.} We can observe that the pruned SNGAN shows the best performance at the sparsity of 10\%-30\%. 
}
\label{fig:sngan_random}
\end{figure}

\begin{figure}[ht!]
\centering
\includegraphics[width=1\columnwidth ]{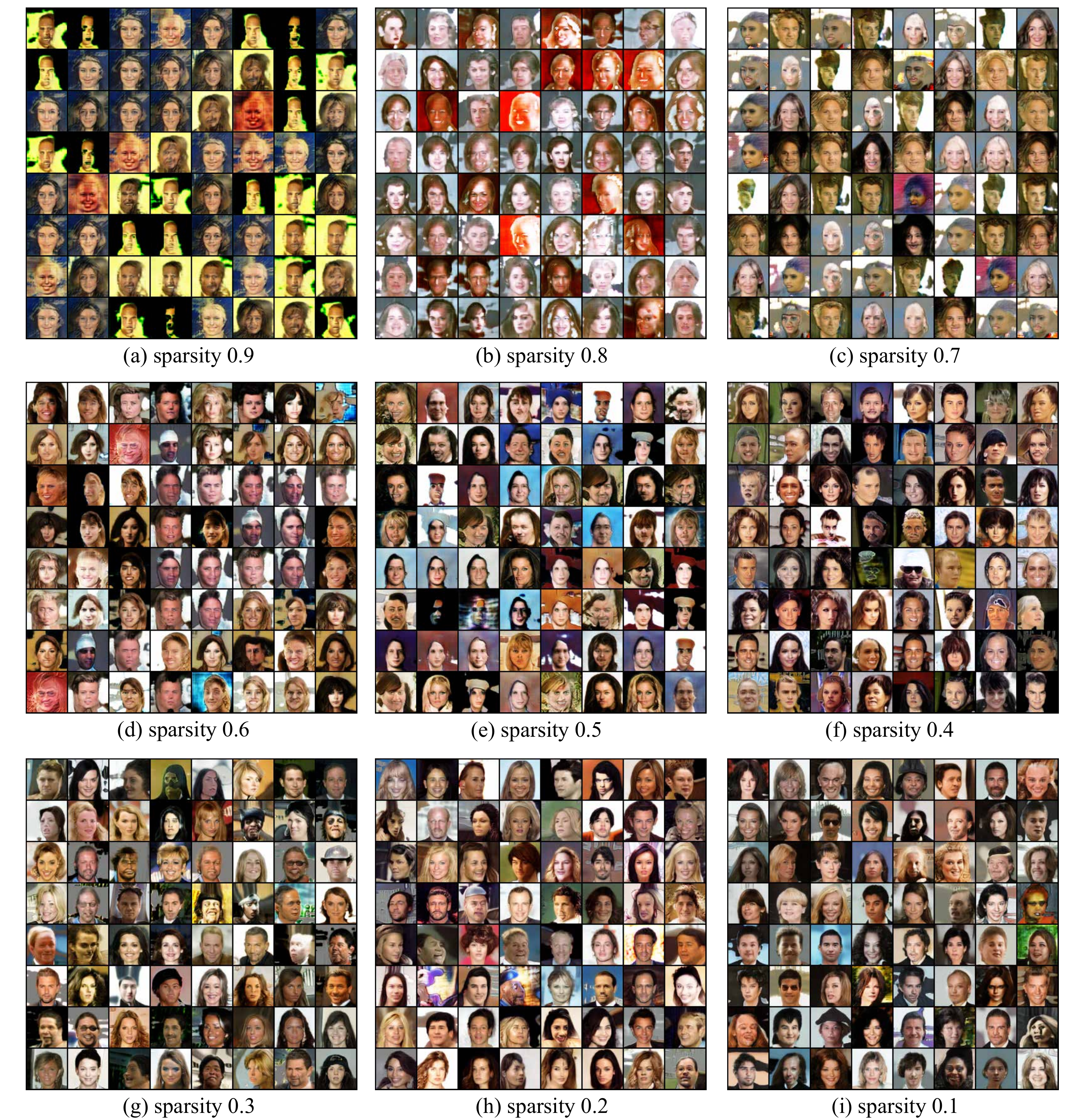}
\caption{ {\bf Qualitative results according to sparsity, pruning a random weight BigGAN on the CelebA dataset.} We can observe that the pruned BigGAN shows the best performance at the sparsity of 10\%-40\%. 
}
\label{fig:biggan_random}
\end{figure}

\begin{figure}[ht!]
\centering
\includegraphics[width=1\columnwidth ]{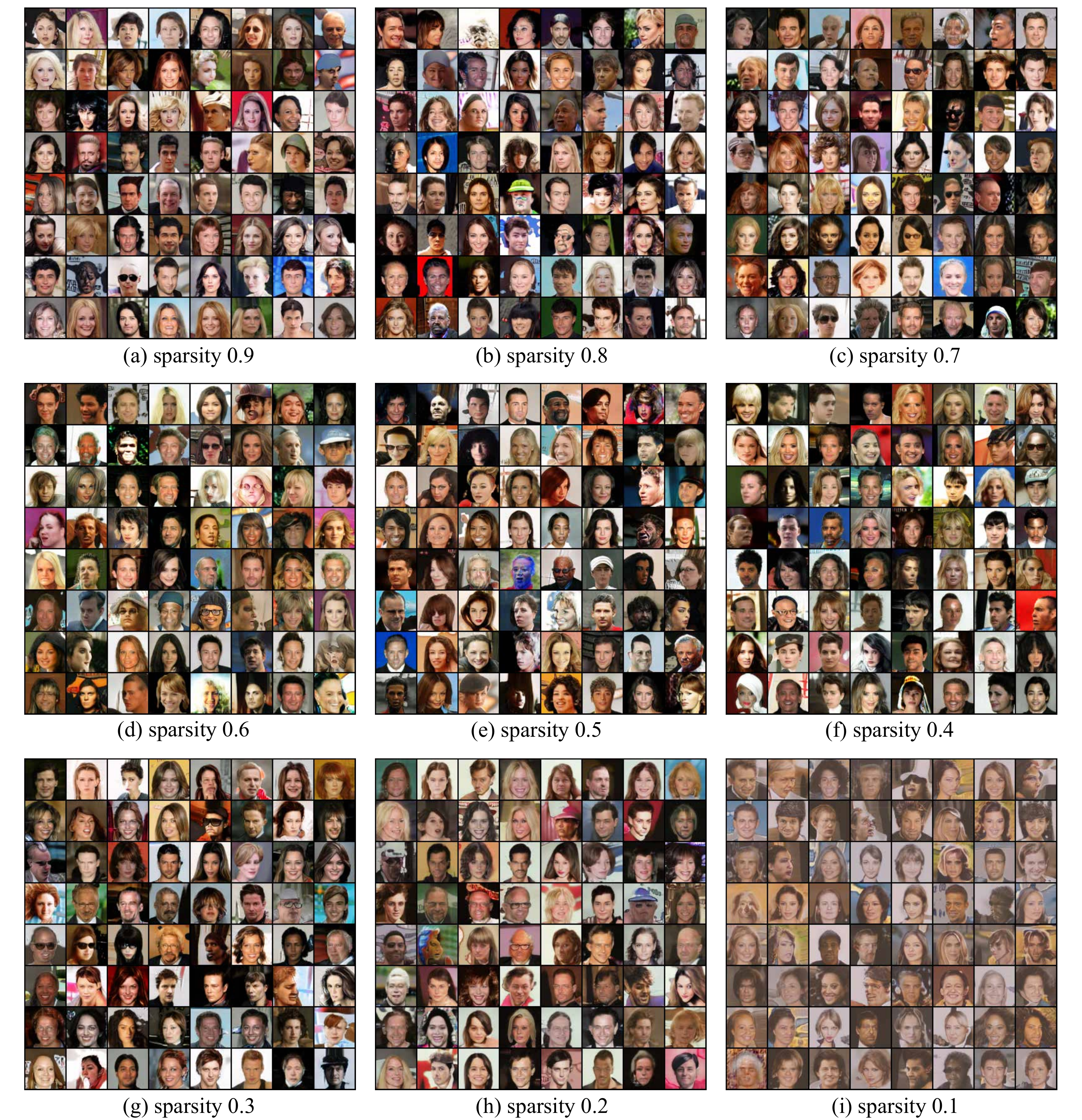}
\caption{ {\bf Qualitative results according to sparsity, pruning a pretrained BigGAN on the CelebA dataset.} We can observe that the pruned BigGAN shows the best performance at the sparsity of 80\%-90\%. 
}
\label{fig:biggan_pretrain}
\end{figure}

\begin{figure}[ht!]
\centering
\includegraphics[width=1\columnwidth ]{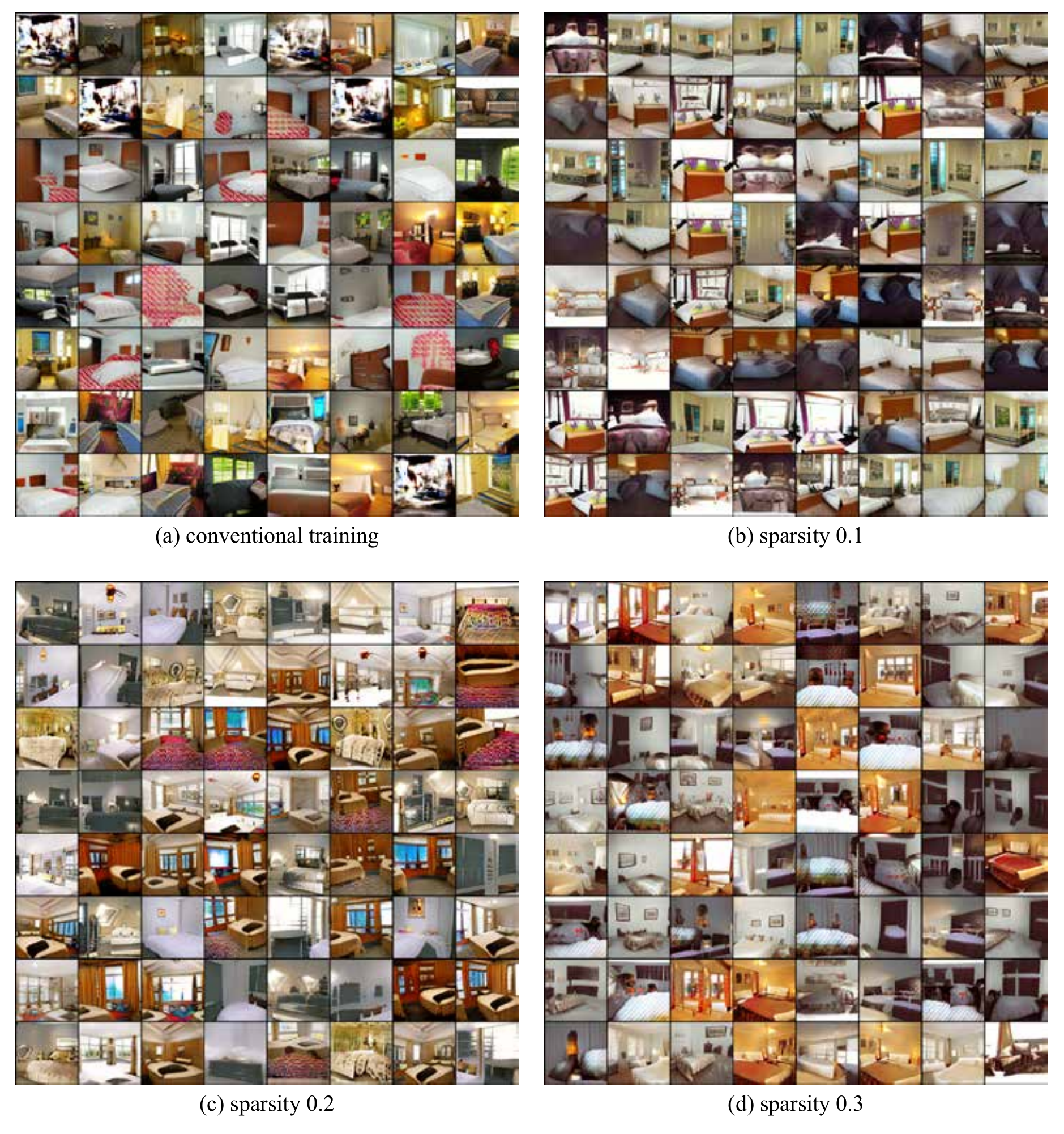}
\caption{ {\bf Qualitative results according to sparsity, pruning a random weight GFMN on the LSUN-Bedroom dataset at 128 resolution.} The pruned GFMN shows the trend that is similar to 64 resolution at 128 resolution.
}
\label{fig:128resolution}
\end{figure}

We provide the images generated by the found subnetwork for various dataset and sparsity ($k$; the portion of weights remained).  
Figure~\ref{Random_LSUN_Bedroom}, Figures~\ref{fig:sngan_random}, Figure~\ref{fig:biggan_random}, and Figure~\ref{fig:biggan_pretrain} show the results for the LSUN-Bedroom and CelebA datasets for randomly initialized GFMN, SNGAN, and BigGAN and for pretrained BigGAN, respectively. Also, we provide higher resolution results on GFMN on LSUN-Bedroom dataset. In Figure~\ref{fig:128resolution}, we apply our a method for 128 resolution and observe that the trend remains the same; there exists SLT in high(er)-capacity generators.

\paragraph{Convergence speed} We obtain a subnetwork from a randomly initialized SNGAN on CelebA by applying Iterative Magnitude Pruning (IMP)~\citep{frankle2018lottery} to compare the convergence speed of the IMP and the our method. When the subnetwork is obtained by IMP with GAN loss~\citep{chen2021gans, kalibhat2021winning}, it fails to achieve good performance in the sparse regime (10\%). To make a comparison in the sparse regime, we run IMP using MMD loss (which stabilizes the training more than GAN loss) and compare it with our method. For 10\% sparsity, ours reaches the same FID (=15) 2.1 times faster (1.7M epochs) than IMP (3.6M epochs) with MMD loss.

\section{Appendix E. Reference of the FID values of other generative models}
For Figure~\ref{fig:random2}, Figure~\ref{fig:pretrain}, Table~\ref{table:epvar_ran}, Table~\ref{table:epvar_pre}, and Figure 10 in the main manuscript, we refer to the FID values of SAGAN, SNGAN, ResNet, DCGAN, TransGAN-XL, and BigGAN that are reported in \citep{li2022revisiting}, \citep{chen2021gans}, \citep{chen2021gans}, \cite{hou2021slimmable}, \citep{jiang2021transgan}, and \citep{kang2022studiogan}, respectively. Note that StudioGAN~\citep{kang2022studiogan} provides an extensive and reliable benchmark score table, reproducing the original papers' performance across various models, datasets, and evaluation metrics.

\end{document}